\documentclass[lettersize,journal]{IEEEtran}
\usepackage{amsmath,amssymb,amsfonts}
\usepackage{algorithmic}
\usepackage{algorithm}
\usepackage{array}
\usepackage[caption=false,font=normalsize,labelfont=sf,textfont=sf]{subfig}
\usepackage{textcomp}
\usepackage{stfloats}
\usepackage{url}
\usepackage{verbatim}
\usepackage{graphicx}
\usepackage{cite}
\usepackage{multirow}
\usepackage{color}
\usepackage{xcolor}
\usepackage{bbding}
\usepackage{booktabs}
\usepackage{hyperref}
\begin{document}

\title{Toward a Multi-View Brain Network Foundation Model: Cross-View Consistency Learning Across Arbitrary Atlases}

\author{Jiaxing Xu, Jingying Ma, Xin Lin, Yuxiao Liu, Kai He, Qika Lin, Yiping Ke,  \\Yang Li, \IEEEmembership{Senior Member, IEEE}, Dinggang Shen, \IEEEmembership{Fellow, IEEE}, Mengling Feng, \IEEEmembership{Senior Member, IEEE}
\thanks{Jiaxing Xu, Jingying Ma, Kai He, Qika Lin, and Mengling Feng are with the Saw Swee Hock School of Public Health at National University of Singapore, Singapore (e-mail:\protect\url{jiaxing.xu@nus.edu.sg}; \protect\url{jingyingma@u.nus.edu}, \protect\url{kai_he@nus.edu.sg}; \protect\url{qikalin@foxmail.com}; \protect\url{ephfm@nus.edu.sg}).}
\thanks{Xin Lin and Yuxiao Liu are with the School of Biomedical Engineering \& State Key Laboratory of Advanced Medical Materials and Devices, ShanghaiTech University, Shanghai 201210, China (e-mail: \protect\url{linxin2024@shanghaitech.edu.cn}; \protect\url{liuyx7@shanghaitech.edu.cn}).}
\thanks{Yiping Ke is with the College of Computing and Data Science, Nanyang Technological University, Singapore (e-mail: \protect\url{ypke@ntu.edu.sg}).}
\thanks{Yang Li is with the School of Automation Science and Electrical Engineering, Beihang University, Beijing 100191, China, and also with the State Key Laboratory of Virtual Reality Technology and Systems, Beihang University, Beijing 100191, China (e-mail:\protect\url{liyang@buaa.edu.cn}).}
\thanks{Dinggang Shen is with the School of Biomedical Engineering \& State Key Laboratory of Advanced Medical Materials and Devices, ShanghaiTech University, Shanghai 201210, China. He is also with Shanghai United Imaging Intelligence, Shanghai 200230, China, and Shanghai Clinical Research and Trial Center, Shanghai 201210, China (e-mail: Dinggang.Shen@gmail.com).}
\thanks{Corresponding author: Dinggang Shen.}}

\markboth{Journal of \LaTeX\ Class Files,~Vol.~14, No.~8, August~2021}%
{Shell \MakeLowercase{\textit{et al.}}: A Sample Article Using IEEEtran.cls for IEEE Journals}


\maketitle

\begin{abstract}
Brain network analysis provides an interpretable framework for characterizing brain organization and has been widely used for neurological disorder identification. Recent advances in self-supervised learning have motivated the development of brain network foundation models. However, existing approaches are often limited by atlas dependency, insufficient exploitation of multiple network views, and weak incorporation of anatomical priors.  In this work, we propose MV-BrainFM, a multi-view brain network foundation model designed to learn generalizable and scalable representations from brain networks constructed with arbitrary atlases. MV-BrainFM explicitly incorporates anatomical distance information into Transformer-based modeling to guide inter-regional interactions, and introduces an unsupervised cross-view consistency learning strategy to align representations from multiple atlases of the same subject in a shared latent space. By jointly enforcing within-view robustness and cross-view alignment during pretraining, the model effectively captures complementary information across heterogeneous network views while remaining atlas-aware. In addition, MV-BrainFM adopts a unified multi-view pretraining paradigm that enables simultaneous learning from multiple datasets and atlases, significantly improving computational efficiency compared to conventional sequential training strategies. The proposed framework also demonstrates strong scalability, consistently benefiting from increasing data diversity while maintaining stable performance across unseen atlas configurations. Extensive experiments on more than 20,000 subjects from 17 fMRI datasets show that MV-BrainFM consistently outperforms 14 existing brain network foundation models and task-specific baselines under both single-atlas and multi-atlas settings. Furthermore, the learned representations exhibit strong anatomical and clinical interpretability, highlighting meaningful brain regions and cross-view consistent patterns associated with neurological conditions. These results suggest that MV-BrainFM provides an effective, efficient, and scalable foundation model for brain network analysis. The source code is available at \url{https://github.com/AngusMonroe/MV-BrainFM}.
\end{abstract}

\begin{IEEEkeywords}
Brain Network, fMRI, Foundation Model, Graph Neural Network, Neurological Disorder
\end{IEEEkeywords}

\section{Introduction}
\IEEEPARstart{B}{rain} network analysis has become a fundamental paradigm in neuroimaging research for characterizing brain organization and for developing biomarkers of neurological and psychiatric disorders. By representing the brain as a graph whose nodes correspond to regions of interest (ROIs) and whose edges encode inter-regional relationships, brain networks provide an interpretable abstraction of complex neuroimaging data~\cite{worsley2002general}. Recent advances in graph neural networks (GNNs) and Transformer-based architectures have significantly improved representation learning for brain networks, leading to strong performance in disease diagnosis, progression modeling, and subject stratification~\cite{li2021braingnn,zhang2022classification,xu2024contrasformer,peng2025biologically,xu2025multiview}.

Despite their progress, most existing brain network models are still trained in a task-specific and atlas-dependent manner, which limits their generalization across datasets, disorders, and parcellation schemes~\cite{wei2026a}. In practice, brain networks are constructed using diverse atlases with varying numbers of ROIs and anatomical granularity~\cite{xu2023data}. Moreover, a single subject can be represented by multiple brain networks derived from different atlases, each capturing complementary aspects of the underlying connectome~\cite{xu2025multi}. Designing models that can effectively leverage such heterogeneous representations remains a major challenge.

\begin{figure*}[t]
\centering
\includegraphics[width=0.85\textwidth]{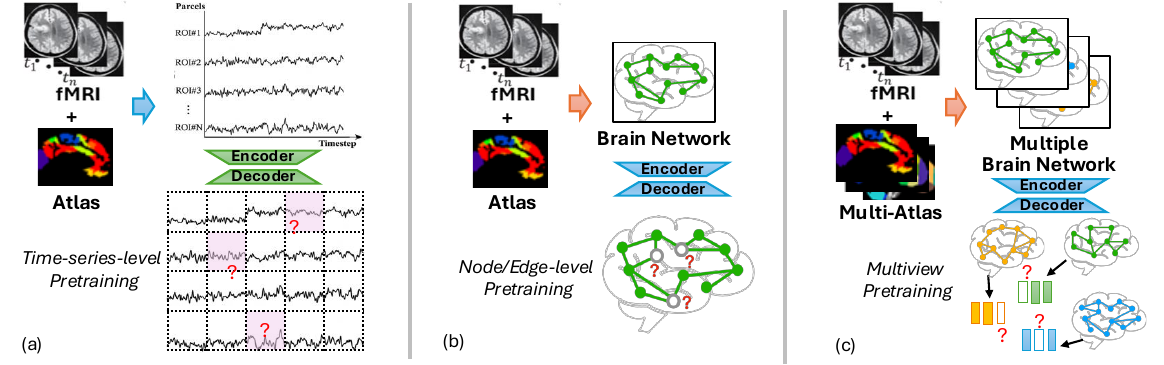}
\small
\begin{tabular}{clccccc}
\hline
\textbf{Category} & \textbf{Model} & \textbf{Venue \& Year} &
\textbf{Graph-Based} &
\textbf{Atlas Dependency} &
\textbf{Multi-View} & \textbf{Training Strategy}\\
\hline
\multirow{2}{*}{(a)}  & BrainLM~\cite{caro2024brainlm} & ICLR 2024 & \XSolidBrush & Single atlas & \XSolidBrush & Sequentially \\
& Brain-JEPA~\cite{dong2024brain} & NeurIPS 2024 & \XSolidBrush & Single atlas & \XSolidBrush & Sequentially \\ \hline
\multirow{4}{*}{(b)}  & PTGB~\cite{yang2023ptgb}  & CHIL 2023 & \Checkmark & Single atlas & \XSolidBrush & Sequentially \\
& BrainMass~\cite{yang2024brainmass} & TMI 2024 & \Checkmark & Single atlas & \XSolidBrush & Simultaneously \\
& GTFM~\cite{wang2026graph} & PR 2026 & \Checkmark & Single atlas & \XSolidBrush & Sequentially \\
& BrainGFM~\cite{wei2026a} & ICLR 2026 & \Checkmark & Multi-atlas  & \XSolidBrush & Sequentially\\ \hline
(c)  & MV-BrainFM (Ours) & 2026 & \Checkmark & Multi-atlas & \Checkmark & Simultaneously\\
\hline
\end{tabular}
\caption{Comparison of representative brain foundation models.}
\label{fig:comparison}
\end{figure*}

Motivated by the success of foundation models in natural language processing~\cite{radford2018improving} and computer vision~\cite{dosovitskiy2020image}, recent studies have begun exploring large-scale self-supervised pretraining for neuroimaging. Early brain foundation models, such as BrainLM~\cite{caro2024brainlm} and Brain-JEPA~\cite{dong2024brain}, focus on fMRI time-series modeling and employ masked reconstruction or latent prediction objectives. While these approaches demonstrate promising transferability, they do not explicitly model brain networks and are inherently tied to fixed ROI definitions and fixed time-series lengths, making them less suitable for connectome-based analysis. More recent graph-based foundation models, including PTGB~\cite{yang2023ptgb}, BrainMass~\cite{yang2024brainmass}, and BrainGFM~\cite{wei2026a}, operate directly on brain networks and offer improved efficiency and interpretability. However, as summarized in Fig.~\ref{fig:comparison}, most existing graph-based foundation models either assume a single atlas during training or treat different atlases independently, without explicitly leveraging multiple views of the same subject.

Another important limitation shared by many existing approaches is the insufficient incorporation of anatomical priors. Standard GNNs~\cite{kipf2017semi,velivckovic2017graph} and Transformer-based models~\cite{vaswani2017attention,yun2019graph} typically rely on learned message passing or attention mechanisms that treat all ROI pairs as equally plausible interaction candidates. This design neglects well-established neurobiological evidence that physical distance between brain regions strongly modulates functional connectivity~\cite{vertes2012simple,ercsey2013predictive,goni2014resting}. Ignoring such spatial priors can lead to biologically implausible interactions and limits the interpretability of learned representations, especially when models are scaled to heterogeneous datasets.

In this work, we propose \textit{MV-BrainFM}, a Multi-View Brain Network Foundation Model that explicitly addresses these limitations. MV-BrainFM is designed to (i) operate on brain networks constructed from arbitrary atlases, (ii) incorporate anatomical distance information as an inductive bias for representation learning, and (iii) leverage cross-view consistency learning to exploit complementary information from multiple brain network views of the same subject. Instead of enforcing hard node correspondence across atlases, MV-BrainFM maps heterogeneous networks into a shared latent space and aligns them at the representation level. A distance-aware attention mechanism is introduced to modulate inter-ROI interactions according to physical proximity, improving spatial plausibility and robustness. During pretraining, an unsupervised cross-view consistency objective further encourages representations from different atlases to be consistent while preserving view-specific characteristics, enabling the model to learn atlas-invariant yet informative subject representations. 

Importantly, the proposed multi-view design enables MV-BrainFM to scale effectively with increasing numbers of atlases and datasets by jointly leveraging heterogeneous brain network views within a unified training paradigm. This strategy avoids the sequential dataset training adopted by many existing brain foundation models and improves computational efficiency while enabling the model to learn more generalizable representations. We evaluate MV-BrainFM on 64K+ brain networks of 20K+ subjects from 17 fMRI datasets spanning different tasks and atlas configurations. Experimental results show that MV-BrainFM consistently outperforms existing brain foundation models and task-specific baselines while maintaining high computational efficiency. In addition, the learned representations exhibit clear anatomical and clinical interpretability, highlighting meaningful ROIs and cross-view consistent patterns associated with disease.

Our main contributions are summarized as follows:
\begin{itemize}
\item We propose MV-BrainFM, a unified multi-view brain network foundation model that supports brain networks constructed from arbitrary atlases and enables scalable multi-atlas representation learning.
\item We introduce a distance-aware modeling framework that explicitly incorporates anatomical spatial priors into Transformer-based brain network learning.
\item We develop a cross-view consistency learning strategy for unsupervised pretraining, enabling effective integration of complementary brain network views and improved scalability with increasing atlas diversity.
\item We demonstrate state-of-the-art performance, strong interpretability, and superior computational efficiency across multiple datasets with various atlas configurations.
\end{itemize}

\section{Related Works}

\subsection{GNNs in Brain Network Analysis}

GNNs have become a dominant tool in brain network analysis due to their ability to model complex topological structures. Early methods like BrainNetCNN~\cite{kawahara2017brainnetcnn} introduced specialized convolutional filters to capture edge-, node-, and graph-level patterns. GCN-based approaches~\cite{ktena2017distance,parisot2018disease} were used to learn pairwise similarities across subjects, while PRGNN~\cite{li2020pooling} and BrainGNN~\cite{li2021braingnn} introduced pooling strategies to enforce group consistency and identify salient ROIs, respectively. MG2G~\cite{xu2021graph} presented a two-stage framework combining self-supervised representation learning with disease classification. 
GAT-LI~\cite{hu2021gat} introduces a graph attention network framework for functional brain network classification and interpretation, leveraging attention mechanisms and GNNExplainer~\cite{ying2019gnnexplainer} to identify important connectivity patterns associated with ASD.
BNT~\cite{kan2022brain} applied a graph Transformer to learn pairwise connection strengths among brain regions across individuals. BioBGT~\cite{peng2025biologically} proposed a node importance encoding technique that captures the structural importance of nodes in global information propagation, highlighting the biological properties of the brain structure. Some works~\cite{bannadabhavi2023community,shehzad2025multiscale} improve the diagnosis of brain diseases by capturing the hierarchical structure of brain connections at different spatial scales. Recent contrastive methods~\cite{xu2024contrastive,xu2024contrasformer} focus on enhancing group differentiation to improve generalization. BQN~\cite{yangwe} incorporates quadratic networks to possess better universal approximation properties and implicitly performs community detection along with representation learning.

\subsection{Brain Network Foundation Models}

The concept of foundation models has recently been extended to neuroimaging, aiming to learn transferable representations from large-scale unlabeled brain data. Early efforts primarily focus on fMRI time-series modeling, where brain activity is treated as a spatiotemporal signal. BrainLM~\cite{caro2024brainlm} introduces masked autoencoding to reconstruct fMRI time-series patches and demonstrates the feasibility of large-scale self-supervised pretraining. Brain-JEPA~\cite{dong2024brain} further advances this direction by adopting a joint-embedding predictive architecture that predicts latent representations instead of raw signals, improving semantic quality and generalization across demographic and clinical tasks. Nevertheless, these time-series-based models reconstruct raw BOLD signals, which can amplify noise and do not explicitly model brain networks, making them less suitable for heterogeneous connectome analysis.

More recent studies have shifted attention toward graph-based foundation models, which operate directly on brain networks and offer improved efficiency and interpretability. Existing graph foundation models aim to learn transferable representations through large-scale self-supervised pretraining on diverse graph datasets. However, these general-purpose frameworks~\cite{hu2020strategies,hou2022graphmae,liu2025graph} are not directly suitable for brain networks, which exhibit unique characteristics such as fixed anatomical topology and atlas-dependent node definitions.
PTGB~\cite{yang2023ptgb} proposes self-supervised pretraining of graph neural networks for brain networks and introduces atlas mapping to enable limited cross-atlas transfer. BrainMass~\cite{yang2024brainmass} leverages large-scale pseudo-functional connectivity and combines masked ROI modeling with latent representation alignment to enhance generalization across datasets. GTFM~\cite{wang2026graph} introduces an encoder-decoder architecture to pretrain the brain network foundation model and integrates node and edge embeddings to extract connectivity features. These models demonstrate strong few-shot and zero-shot performance but typically assume a single atlas during training and do not explicitly exploit multiple parcellations of the same subject as complementary views.
BrainGFM~\cite{wei2026a} represents a notable step toward general-purpose brain network foundation modeling by introducing prompt tuning and pretraining across multiple disorders and atlases. While BrainGFM supports multiple atlases, it primarily treats different atlases independently and does not explicitly enforce cross-view consistency among them. Moreover, anatomical priors such as inter-ROI spatial distance are only implicitly encoded or omitted in most existing graph-based approaches, limiting spatial interpretability.

In contrast to prior work, our approach explicitly targets multi-view brain network foundation modeling by jointly leveraging multiple atlases of the same subject through unsupervised cross-view consistency learning. Moreover, by integrating anatomical distance priors and supporting arbitrary atlas configurations within a unified framework, our method addresses key limitations of existing time-series-based and graph-based foundation models.

\section{Proposed Method}

\begin{figure*}[t]
\centering
\includegraphics[width=.9\textwidth]{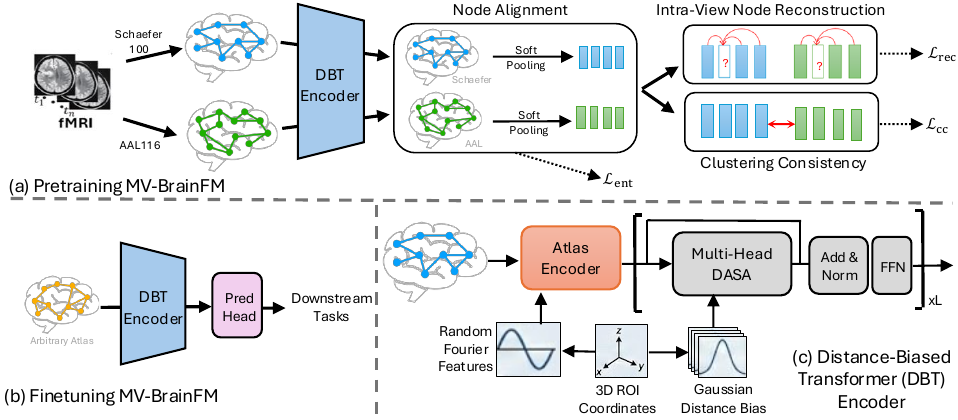}
\caption{The architecture of our proposed MV-BrainFM using Schaefer100 and AAL116 as example. (a) Multiple brain network datasets with various atlases are utilized for pretraining. (b) Datasets with arbitrary atlases can be introduced to finetune the model for downstream tasks with a simple prediction head. (c) The proposed Distance-Biased Transformer (DBT) Encoder.}
\label{fig:framework}
\end{figure*}

Fig.~\ref{fig:framework} illustrates the overall pipeline of MV-BrainFM. Given a brain network constructed from an arbitrary atlas, it is encoded by a Distance-Biased Transformer (DBT) Encoder, which includes (1) a Random Fourier Atlas Encoder (Section~\ref{sec:atlas_rff}) to transform the input connectivity matrix into node embeddings by injecting spatial coordinate priors for heterogeneous ROI sets across atlases, and (2) a Distance-Aware Self-Attention (Section~\ref{sec:distance_biased_attention}) to encourage each attention head to focus on interactions at specific anatomical distance regimes. 
To exploit complementary information from multiple atlas-specific views, we further introduce an unsupervised cross-view consistency pretraining strategy (Section~\ref{sec:cross_view_consistency}). Specifically, node representations from different atlases are aligned into a shared supernode space via a learnable node alignment module, optimized with intra-view masked node reconstruction and clustering-based cross-view consistency. After pretraining, the backbone can be efficiently adapted to downstream clinical tasks using a lightweight linear prediction head. 

\subsection{Random Fourier Atlas Encoder}
\label{sec:atlas_rff}

A key challenge in large-scale brain network modeling lies in handling networks constructed from heterogeneous brain atlases, which may differ substantially in the number of ROIs, spatial resolution, and anatomical granularity. To address this issue, we propose a Random Fourier Atlas Encoder that maps ROI-level spatial coordinates from arbitrary atlases into a unified latent space while explicitly preserving atlas identity.

Let a brain network be represented as $\mathcal{G} = (\mathcal{V}, \mathcal{E})$, where each node $v_i \in \mathcal{V}$ corresponds to an ROI defined under a specific atlas. For atlas $a$, we denote the number of ROIs as $N_a=|\mathcal{V}_a|$, and associate each ROI with a fixed 3D anatomical coordinate $\mathbf{c}_i \in \mathbb{R}^3$ in a common stereotaxic space (e.g., MNI space~\cite{fonov2009unbiased}). The input connectivity matrix computed by Pearson's correlation for atlas $a$ is given by $\mathbf{X}^{a} \in \mathbb{R}^{N_a \times N_a}$.

To accommodate varying atlas resolutions, we employ atlas-specific linear projections that map input features into a shared embedding dimension $d$:
\begin{equation}
\mathbf{H}^{a} = \mathbf{X}^{a} \mathbf{W}^{a}, 
\quad \mathbf{W}^{a} \in \mathbb{R}^{N_a \times d},
\end{equation}
where each atlas has its own learnable projection matrix, enabling the model to respect atlas-dependent feature statistics while producing embeddings in a common latent space.

To encode anatomical spatial information in a manner that is invariant to atlas resolution yet sensitive to inter-ROI geometry, we adopt Random Fourier Features (RFF) to approximate a shift-invariant kernel over 3D coordinates. Specifically, we sample a set of $K$ learnable frequency vectors
\begin{equation}
\mathcal{Z} = \{\mathbf{z}_k\}_{k=1}^{K}, \quad \mathbf{z}_k \in \mathbb{R}^3,
\label{eq:freq_vec}
\end{equation}
and compute the Fourier embedding for each ROI coordinate $\mathbf{c}_i$ as:
\begin{equation}
\begin{split}
\phi(\mathbf{c}_i) =
[
\sin(2\pi \mathbf{z}_1^\top \mathbf{c}_i), \cos(2\pi \mathbf{z}_1^\top \mathbf{c}_i), \ldots,\\
\sin(2\pi \mathbf{z}_K^\top \mathbf{c}_i), \cos(2\pi \mathbf{z}_K^\top \mathbf{c}_i)
]
\in \mathbb{R}^{2K}.
\end{split}
\end{equation}
This formulation enables the model to capture multi-scale spatial variations and approximate continuous spatial similarity through a finite-dimensional embedding.

The resulting Fourier features are further projected into the model dimension and combined with the node embedding via residual addition:
\begin{equation}
\tilde{\mathbf{H}}^{a} = \mathbf{H}^{a} + \boldsymbol{\phi}(\mathbf{C}^{a}) \mathbf{W}_{\text{proj}},
\end{equation}
where $\mathbf{C}^{a} \in \mathbb{R}^{N_a \times 3}$ denotes the coordinate matrix for atlas $a$ and $\mathbf{W}_{\text{proj}} \in \mathbb{R}^{2K \times d}$ denotes a learnable parameter matrix. This additive formulation ensures that spatial priors act as an inductive bias rather than dominating learned functional representations.

Unlike atlas-agnostic positional encodings that assume a fixed node set, the proposed encoder explicitly conditions on atlas identity while sharing a common Fourier-based coordinate embedding mechanism across atlases. As a result, it enables a single foundation model to ingest brain networks constructed from arbitrary atlases, facilitating cross-atlas generalization and scalable multi-view pretraining without requiring atlas-specific architectural redesign.

\subsection{Distance-Aware Self-Attention}
\label{sec:distance_biased_attention}

Standard self-attention mechanisms treat all ROI pairs as equally plausible interaction candidates~\cite{vaswani2017attention} and do not explicitly incorporate anatomical priors. However, extensive neurobiological evidence~\cite{vertes2012simple,ercsey2013predictive,goni2014resting} suggests that both structural and functional connectivity are strongly modulated by physical distance between brain regions. To incorporate this prior in a principled and scalable manner, we introduce a distance-aware attention bias through a multi-atlas coordinate encoder.

For each atlas $a$ with $N_a$ ROIs, we precompute a fixed pairwise Euclidean distance matrix
\begin{equation}
\mathbf{D}^{a} \in \mathbb{R}^{N_a \times N_a}, 
\quad
\mathbf{D}^{a}_{ij} = \lVert \mathbf{c}_i^a - \mathbf{c}_j^a \rVert_2,
\end{equation}
where $\mathbf{c}_i^a \in \mathbb{R}^3$ denotes the anatomical coordinate of ROI $i$ for atlas $a$. Since all atlases are defined in the same coordinate system, these distances are directly comparable across atlases and remain fixed during training.

We then parameterize the spatial prior as a \emph{per-head Gaussian attention bias}. For each attention head, we define the attention bias between ROIs $i$ and $j$ as:
\begin{equation}
b_{head}^{a}(i,j)
=
\alpha_{head}
\left(
-
\frac{\left(\mathbf{D}^{a}_{ij} - \mu_{head}^{a}\right)^2}{2 (\sigma_{head}^{a})^2}
\right)
+
\beta_{head},
\end{equation}
where $\alpha_{head} > 0$ is a learnable scaling factor, $\beta_{head}$ is a learnable offset, and $(\mu_{head}^{a}, \sigma_{head}^{a})$ control the preferred distance regime and bandwidth of $head$. This formulation enables each attention head to specialize in modeling interactions at a specific anatomical scale (e.g., short-range or long-range connectivity).

To ensure consistent behavior across atlases with different spatial extents and ROI resolutions, we learn the Gaussian centers and bandwidths in normalized distance units. Specifically, let $dis_{\max}^{a}$ denote the maximum pairwise distance within atlas $a$. We parameterize
\begin{equation}
\mu_{head}^{a} = \tilde{\mu}_{head} \, dis_{\max}^{a},
\qquad
\sigma_{head}^{a} = \tilde{\sigma}_{head} \, dis_{\max}^{a},
\end{equation}
where $\tilde{\mu}_{head} \in (0,1)$ and $\tilde{\sigma}_{head} > 0$ are atlas-independent learnable parameters shared across all atlases. This design ensures that the same attention head corresponds to similar relative spatial regimes across different atlases, enabling coherent multi-atlas modeling without requiring atlas-specific parameterization.

The distance-based attention bias $\mathbf{B}^{a}_{head} \in \mathbb{R}^{N_a \times N_a}$ is symmetric and shared across subjects for a given atlas. Given input node embeddings $\tilde{\mathbf{H}}^a \in \mathbb{R}^{N_a \times d}$, the distance bias is added to the scaled dot-product attention logits before softmax:
\begin{equation}
\mathbf{A}_{head}^{a}=
\mathrm{Softmax}\left(
\frac{\mathbf{Q}_{head}^{a}\mathbf{K}_{head}^{a\top}}{\sqrt{d_{head}}}
+
\mathbf{B}_{head}^{a}
\right), 
\end{equation}
for each head $head=1,\dots,N_{head}$, where $N_{head}$ is the number of attention heads, and $\mathbf{Q}^a_{head}, \mathbf{K}^a_{head} \in \mathbb{R}^{N_a \times d_{head}}$ are linear projections of $\tilde{\mathbf{H}}^a$. The outputs from all heads are concatenated and projected to form the multi-head attention representation:
\begin{equation}
\small
\operatorname{MultiHead}(\tilde{\mathbf{H}}^{a}) =
\mathrm{Concat}(\mathbf{A}^{a}_{1} \mathbf{V}^{a}_{1},\dots,\mathbf{A}^{a}_{N_{head}} \mathbf{V}^{a}_{N_{head}})
\mathbf{W}^{O},
\end{equation}
where $\mathbf{W}^{O} \in \mathbb{R}^{N_{head} d_{head} \times d}$ is the output projection matrix, and $\mathbf{V}^a_{head} \in \mathbb{R}^{N_a \times d_{head}}$ are linear projections of $\tilde{\mathbf{H}}^a$. 

By introducing a head-specific Gaussian bias learned in normalized units, the proposed coordinate encoder provides a lightweight yet expressive mechanism for incorporating anatomical distance priors. Importantly, it naturally supports multiple atlases within a unified parameter space, making it particularly suitable for large-scale multi-atlas foundation model pretraining.

The output of distance-aware self-attention (DASA) is computed by a gated residual update:
\begin{equation}
\small
\text{DASA}(\tilde{\mathbf{H}}^{a}) = \tilde{\mathbf{H}}^{a} + \text{LN}(\gamma(\tilde{\mathbf{H}}^{a}) \odot \operatorname{MultiHead}(\tilde{\mathbf{H}}^{a})),
\end{equation}
where $\gamma(\cdot)$ is a learnable sigmoid gate, $\odot$ denotes the element-wise multiplication, and $\text{LN}(\cdot)$ denotes a layer normalization.

Finally, the $l$-th Transformer layer with distance-biased self-attention is computed as:
\begin{equation}
\tilde{\mathbf{H}}^{(l+1)}=
\text{LN}\Big(
\tilde{\mathbf{H}}^{(l)}+
\text{FFN}\big(
\text{DASA}^{(l)}(\tilde{\mathbf{H}}^{(l)})
\big)
\Big),
\end{equation}
where $\text{FFN}(\cdot)$ denotes a feed-forward network.

\subsection{Cross-View Consistency Pretraining}
\label{sec:cross_view_consistency}

Brain networks derived from different construction pipelines (e.g., atlas with various number of ROIs) provide complementary views of the same underlying subject-specific connectome~\cite{liu2023deep,wang2024multiview,xu2025multi}. However, naively pooling multi-view data can cause view-specific shortcuts and limit generalization. We therefore pretrain the proposed foundation model using an unsupervised cross-view consistency objective that encourages representations from different views of the same subject to agree in a shared latent space, while retaining sufficient flexibility to model view-dependent information.

For each subject, we assume up to $M$ available views $\{\mathcal{G}^{a_m}\}_{m=1}^{M}$, where each view corresponds to a brain network constructed under a specific atlas. Let the DBT Encoder $f_{\theta}(\cdot)$ (Sections~\ref{sec:atlas_rff} and \ref{sec:distance_biased_attention}) produce node-level embeddings $\tilde{\mathbf{H}}^{a_m} = f_{\theta}\!\left(\mathcal{G}^{a_m}\right) \in \mathbb{R}^{N_{a_m} \times d}$. 

Since $N_{a_m}$ may differ across views, direct alignment across nodes is ill-posed.
To enable cross-view comparison, we introduce a learnable Node Alignment module that maps each view into a shared latent space with a fixed number of $N_{super}$ supernodes. Concretely, given $\tilde{\mathbf{H}}^{a_m}$, we compute (i) transformed node features and (ii) soft assignment scores:
\begin{equation}
\mathbf{F}^{a_m} = \text{GCN}_{\text{feat}}\!\left(\tilde{\mathbf{H}}^{a_m}, \mathbf{Z}^{a_m}\right) \in \mathbb{R}^{N_{a_m} \times d}, 
\end{equation}
\begin{equation}
\small
\mathbf{S}^{a_m} = \text{Softmax}\!\left(\text{GCN}_{\text{pool}}\!\left(\tilde{\mathbf{H}}^{a_m}, \mathbf{Z}^{a_m}\right)\right) \in \mathbb{R}^{N_{a_m} \times N_{super}},
\label{eq:supernode}
\end{equation}
where the adjacency matrix $\mathbf{Z}^{a_m} \in \mathbb{R}^{N_{a_m} \times N_{a_m}}$ is computed by thresholding the input connectivity matrix $\mathbf{X}^{a_m}$. The threshold is set to 0.3 as same as in the existing works~\cite{li2021braingnn,wei2026a}. The aligned supernode representation for view $a_m$ is obtained by soft pooling~\cite{ying2018hierarchical}:
\begin{equation}
\hat{\mathbf{H}}^{a_m} = \mathbf{S}^{a_m\top} \mathbf{F}^{a_m} \in \mathbb{R}^{N_{super} \times d}.
\end{equation}

\noindent
\paragraph{Entropy Regularization for Stable Node Alignment}
The soft assignment matrix $\mathbf{S}^{a_m}$ defines how nodes in the view $a_m$ are mapped to the $N_{super}$ shared supernodes. Without additional constraints, the assignment may become overly confident early in training, leading to degenerated solutions where only a few latent nodes are utilized. To mitigate this issue, we introduce an entropy-based regularization term that explicitly encourages high-entropy, non-collapsed assignments.

Concretely, we compute the categorical entropy of the assignment distribution for each input node and average it across nodes and views:
\begin{equation}
\mathcal{L}_{\text{ent}} =
\frac{1}{M}\sum_{m=1}^{M}
\frac{1}{N_{a_m}}
\sum_{i=1}^{N_{a_m}}
\mathcal{H}\!\left(\mathbf{S}^{a_m}_{i}\right),
\end{equation}
where $\mathbf{S}^{a_m}_{i}$ denotes the assignment probabilities of node $i$ in view $a_m$ over the $N_{super}$ supernodes, and $\mathcal{H}(\cdot)$ is the Shannon entropy. This term penalizes prematurely peaked assignments and encourages the alignment module to explore diverse latent-node associations.

By maintaining sufficiently diffuse assignments during early training, the entropy regularization improves optimization stability and prevents trivial alignment patterns, while still allowing sharper, more discriminative assignments to emerge as training progresses.

\paragraph{Intra-View Node Reconstruction}
Before imposing cross-view consistency, we first strengthen the quality and robustness of node representations within each view through a reconstruction objective. The motivation is to ensure that each view-specific encoder learns informative and stable node embeddings, rather than relying on spurious correlations that may harm subsequent cross-view alignment.

Concretely, for each available view $a_m$, we obtain its node embeddings $\hat{\mathbf{H}}^{a_m} \in \mathbb{R}^{N_{super} \times d}$. We then sample a random mask and split indices into a kept set $\mathcal{I}_{\text{keep}}$ and a masked set $\mathcal{I}_{\text{mask}}$ with $\mathcal{I}_{\text{keep}} \cup \mathcal{I}_{\text{mask}}=\{1,\ldots,N_{super}\}$ and $\mathcal{I}_{\text{keep}} \cap \mathcal{I}_{\text{mask}}=\emptyset$. The model is provided only the kept embeddings $\hat{\mathbf{H}}^{a_m}_{\mathcal{I}_{\text{keep}}}$, while the masked embeddings are replaced by a mask token (implemented as zeros), yielding an input sequence $\mathbf{Q}^{a_m}$. A lightweight Transformer decoder $g_{\phi}^{a_m}(\cdot)$ is then applied to predict the masked node representations:
\begin{equation}
\hat{\mathbf{Q}}^{ a_m} = g_{\phi}^{a_m}\!\left(\mathbf{Q}^{a_m}\right) \in \mathbb{R}^{N_{a_m} \times d}.
\end{equation}
The reconstruction loss is computed only on masked nodes using a Smooth-$L_1$ objective:
\begin{equation}
\mathcal{L}_{\text{rec}}
=
\frac{1}{M}
\sum_{m=1}^{M}
\frac{1}{|\mathcal{I}_{\text{mask}}|}
\sum_{i \in \mathcal{I}_{\text{mask}}}
\left\|
\hat{\mathbf{Q}}^{a_m}_{i} - \hat{\mathbf{H}}^{a_m}_{i}
\right\|_{L_1}.
\end{equation}
By requiring the model to recover masked node embeddings from their unmasked context, this objective encourages each view to learn robust node representations that capture meaningful intra-view structure. This within-view robustness is particularly important in our setting with heterogeneous views (e.g., varying atlas resolutions and connectivity definitions), and serves as a stable foundation before enforcing cross-view clustering consistency.

\paragraph{Clustering Consistency Loss}
Given the aligned supernode embeddings from all available views, we enforce cross-view agreement via a clustering-based consistency objective. Let $\mathbf{T} \in \mathbb{R}^{M \times N_{super} \times d}$ denote the aligned embeddings, where $M$ is the number of views, and $d$ is the feature dimension. For each view-supernode pair $(m,q)$, we project $\mathbf{T}_{m,q}$ to $P$ prototypes using a linear mapping:
\begin{equation}
\mathbf{p}_{m,q} = \mathrm{Softmax}\!\left(\text{MLP}_{proto}(\mathbf{T}_{m,q})\right) \in \mathbb{R}^P.
\label{eq:prototype}
\end{equation}

To encourage view-invariant representations, we require the prototype assignment distributions $\{\mathbf{p}_{m,q}\}_{m=1}^{M}$ to be consistent across views for each supernode $q$. We first compute the mean assignment distribution across views:
\begin{equation}
\bar{\mathbf{p}}_{q} = \frac{1}{M} \sum_{m=1}^{M} \mathbf{p}_{m,q}.
\end{equation}
We then minimize the symmetric Kullback-Leibler (KL) divergence $D_{\mathrm{KL}}(\cdot\,\|\,\cdot)$ between each view-specific distribution and the mean distribution as a clustering consistency loss:
\begin{equation}
\small
\mathcal{L}_{\text{cc}}
=
\frac{1}{2MN_{super}}
\sum_{m=1}^{M}\sum_{q=1}^{N_{super}}
\left[
D_{\mathrm{KL}}\!\left(\mathbf{p}_{m,q}\,\|\,\bar{\mathbf{p}}_{q}\right)
+
D_{\mathrm{KL}}\!\left(\bar{\mathbf{p}}_{q}\,\|\,\mathbf{p}_{m,q}\right)
\right].
\end{equation}

This objective enforces that different views of the same subject yield similar prototype assignment patterns at the aligned supernode level, thereby promoting cross-view invariance while remaining fully unsupervised.

\paragraph{Overall Pretraining Objective}
The final pretraining objective combines within-view robustness, cross-view alignment, and stable node assignment regularization:
\begin{equation}
\mathcal{L}_{\text{pretrain}}
=
\mathcal{L}_{\text{rec}}
+ \mathcal{L}_{\text{cc}}
+ \mathcal{L}_{\text{ent}}.
\end{equation}
Here, $\mathcal{L}_{\text{rec}}$ enforces intra-view node reconstruction, $\mathcal{L}_{\text{cc}}$ promotes cross-view clustering consistency in the shared latent space, and $\mathcal{L}_{\text{ent}}$ regularizes the node alignment module. In practice, the cross-view consistency term $\mathcal{L}_{\text{cc}}$ is applied only when at least two views are available in a mini-batch.

Existing multi-atlas brain network foundation models~\cite{yang2023ptgb,wei2026a} typically pretrain sequentially across datasets constructed with different atlases, where the model is optimized on one dataset before moving to the next. Such a training strategy may introduce dataset-order sensitivity and limit the model’s ability to jointly capture complementary information across atlases. In contrast, MV-BrainFM adopts a unified multi-view pretraining strategy in which mini-batches from all datasets are shuffled and jointly sampled within each training epoch. Multiple atlas-specific views of the same subject are optimized simultaneously within a mini-batch under the proposed consistency objective. This design enables the model to learn atlas-invariant representations in a single training process, mitigating order effects across both datasets and views.

\paragraph{Fine-Tuning for Downstream Tasks}
After unsupervised pretraining, we adapt the backbone encoder to downstream classification tasks using a lightweight linear prediction head optimized with the standard cross-entropy loss. Our model takes single atlas brain networks for fintuning without relying on multiview input. During fine-tuning, the pretrained encoder parameters are updated jointly with the task-specific head. This protocol allows MV-BrainFM to effectively transfer view-invariant and anatomically informed representations learned from heterogeneous multi-view data to label-scarce clinical settings.

\section{Experiments}

\subsection{Datasets}

We train and evaluate MV-BrainFM on 20,786 subjects from 13 publicly available and 4 private functional brain network datasets. In total, 64,398 brain networks are constructed using six different atlases (Schaefer100, Schaefer200, Schaefer500~\cite{schaefer2018local}, AAL116~\cite{tzourio2002automated}, Craddock200~\cite{craddock2012whole}, and BASC122~\cite{BELLEC20101126}). The 13 public datasets were previously collected under appropriate ethical approvals, with informed consent obtained by the respective data collection institutions. The 4 private datasets (Renji, SMHC, DAMOMRI, and Huashan) were collected under institutional review board approval, and written informed consent was obtained from all participants by the original data providers. Dataset statistics are summarized in Table~\ref{tab:dataset_statistic}.

\begin{table*}[t]
\centering
\caption{Statistics and class information of brain network datasets used in this work. Renji, SMHC, DAMOMRI, and Huashan are private datasets, while the rest are publicly available.}
\begin{tabular}{ccccccc}
\hline
                            & Datasets  & Sex   (F/M) & Age   (mean $\pm$ std) & Class\# & Subjects\# & Atlases                                         \\ \hline
\multirow{12}{*}{Pretrain}  & HBN~\cite{alexander2017open}       & NA          & NA                 & -       & 12427      & Schaefer200,   AAL116, Craddock200              \\
  & ADHD~\cite{adhd2012adhd}      &       383/230      &         12.1 $\pm$   3.3           & -       & 613        & Schaefer100,   Schaefer200, Schaefer500         \\
                            & ADNI~\cite{dadi2019benchmarking}      &   610/552          &        71.8 $\pm$   6.9            & -       & 1162       & Schaefer100,   Schaefer200, Schaefer500, AAL116         \\
                            & ABIDE~\cite{craddock2013neuro}     &    122/697         &           16.9 $\pm$   7.9         & -       & 819        & Schaefer100,   Schaefer200, Schaefer500, AAL116 \\
                            & PPMI~\cite{marek2011parkinson}      & 82/127      & 62.9 $\pm$   9.5       & -       & 209        & Schaefer100,   Schaefer200, Schaefer500, AAL116 \\
                            & Taowu~\cite{marek2011parkinson}     & 17/23       & 65.0 $\pm$   5.0       & -       & 40         & Schaefer100,   Schaefer200, Schaefer500, AAL116 \\
                            & Neurocon~\cite{marek2011parkinson}  & 22/19       & 68.0 $\pm$   11.0      & -       & 41         & Schaefer100,   Schaefer200, Schaefer500, AAL116 \\
                            & OASIS~\cite{marcus2007open}     & 115/135     & 72.4   $\pm$ 4.9       & -       & 250        & Schaefer100,   Schaefer200, Schaefer500         \\
                            & Renji~\cite{liu2022multiscale}     & 69/228      & 65.3  $\pm$ 7.2        & -       & 297        & Schaefer100,   Schaefer200, Schaefer500         \\
                            & SMHC      & NA            & NA                    & -       & 122        & Schaefer100,   Schaefer200, Schaefer500         \\
                            & REST-meta-MDD~\cite{yan2019reduced}       &     438/298        &         33.7   $\pm$ 11.1           & -       & 736        & AAL116,   Craddock200                           \\
                            & DAMOMRI      & NA          & NA                 & -       & 471        & Schaefer100,   Schaefer200, Schaefer500, AAL116 \\ \hline 
\multirow{6}{*}{Downstream}             & ADHD~\cite{adhd2012adhd}      & 478/289     & 12.0   $\pm$ 3.2       & 4       & 768        & Schaefer100,   Schaefer200                      \\
& ADNI~\cite{dadi2019benchmarking}      &     754/697        &      72.5 $\pm$   7.1              & 5       & 1455       & Schaefer100,   Schaefer200, AAL116, BASC122         \\
& ABIDE~\cite{craddock2013neuro}     & 152/873     & 16.5 $\pm$   7.4       & 2       & 1025       & Schaefer100,   Schaefer200, AAL116, BASC122                      \\
                            & HCP-Gender~\cite{van2013wu} &     495/585        &             NA       & 2       & 1080       & Schaefer100,   Schaefer200, AAL116, BASC122                      \\
                            & HCP-Age~\cite{van2013wu}    &    495/585         &        NA            & 4       & 1080       & Schaefer100,   Schaefer200, AAL116, BASC122                      \\
                            & Huashan~\cite{ding2016progression}   & 495/290     & 65.0   $\pm$ 7.7       & 3       & 785        & Schaefer100,   Schaefer200                      \\ \hline 
\end{tabular}
\begin{flushleft}
$^{\dagger}$ NA: demographic information was not publicly available for these datasets.
\end{flushleft}
\label{tab:dataset_statistic}
\vspace{-0.2cm}
\end{table*}

\noindent
\textbf{HBN.}~\cite{alexander2017open} The Healthy Brain Network (HBN) dataset provides neuroimaging and behavioral data from children and adolescents, aimed at understanding brain development and disorders in youth. The dataset includes fMRI, structural MRI, and other neuroimaging modalities, along with a wide range of behavioral and clinical measures.

\noindent
\textbf{ADHD.}~\cite{adhd2012adhd} The ADHD dataset includes fMRI scans from individuals diagnosed with ADHD and healthy controls, collected across multiple sites, to explore the neurobiological basis of ADHD.

\noindent
\textbf{ADNI.}~\cite{dadi2019benchmarking} The Alzheimer’s Disease Neuroimaging Initiative (ADNI) aims to evaluate whether MRI, PET, biological markers, and clinical/neuropsychological assessments can be combined to track the progression of mild cognitive impairment (MCI) and early Alzheimer’s Disease (AD). Raw images were obtained from the ADNI database (\url{adni.loni.usc.edu}); further details are available at \url{http://www.adni-info.org}.

\noindent
\textbf{ABIDE.}~\cite{craddock2013neuro} The Autism Brain Imaging Data Exchange (ABIDE) aggregates functional brain imaging data from multiple international sites to support autism spectrum disorder (ASD) research. ASD is characterized by stereotyped behaviors and symptoms such as irritability, hyperactivity, depression, and anxiety. Subjects are classified into two groups: Typical Control (TC) and individuals with ASD.

\noindent
\textbf{PPMI.}~\cite{marek2011parkinson} The Parkinson’s Progression Markers Initiative (PPMI) is an ongoing, multi-site study designed to identify biomarkers associated with PD risk, onset, and progression. 

\noindent
\textbf{Taowu and Neurocon.}~\cite{badea2017exploring} The Taowu and Neurocon datasets, released by ICI, are among the earliest image datasets made available for Parkinson’s research. These datasets comprise age-matched subjects collected from a single machine and site, encompassing both normal control individuals and patients diagnosed with PD.

\noindent
\textbf{OASIS.}~\cite{marcus2007open} The Open Access Series of Imaging Studies (OASIS) includes neuroimaging data from individuals with MCI, AD, and healthy controls, providing a resource for studying aging and neurodegenerative diseases. 

\noindent
\textbf{Renji.}~\cite{liu2022multiscale} The Renji dataset includes fMRI data from individuals with vascular cognitive impairment (VCI) and healthy controls, collected at Renji Hospital in Shanghai, to study the impact of vascular disorders on brain connectivity.

\noindent
\textbf{SMHC.} The SMHC dataset contains resting-state fMRI data from individuals diagnosed with melancholic and atypical depression, collected at the Shanghai Mental Health Center.

\noindent
\textbf{REST-meta-MDD.}~\cite{yan2019reduced} The REST-meta-MDD dataset is part of the Meta-MDD project from the DIRECT Consortium, containing neuroimaging data from individuals with major depressive disorder (MDD) to investigate brain alterations associated with depression.

\noindent 
\textbf{DAMOMRI.} Drug Addiction Multi-Organ MRI Dataset (DAMOMRI) is a dataset collected by the Yunnan Technological Innovation Center of Drug Addiction Medicine, Yunnan University, which includes multi-organ MRI data of individuals with substance use disorders and healthy controls, covering the brain, heart, liver, and kidneys, together with anxiety, depression, cognitive, and related assessments.

\noindent
\textbf{HCP-Gender and HCP-Age.}~\cite{van2013wu} Human Connectome Project (HCP) dataset, which is a comprehensive publicly available neuroimaging dataset that includes both imaging data and a wide range of behavioral and cognitive data.

\noindent
\textbf{Huashan.}~\cite{ding2016progression}
The Huashan dataset includes fMRI data from individuals diagnosed with MCI and AD, as well as healthy controls, collected at Huashan Hospital in Shanghai, contributing to research on early neurodegeneration in MCI and AD.

The functional brain networks of ABIDE, PPMI, Taowu, and Neurocon are constructed by Xu et al.~\cite{xu2023data}, while the ones of HCP-Gender and HCP-Age are constructed by following the settings in Said et al.~\cite{said2023neurograph}. For the ADNI, ADHD, OASIS, Renji, and Huashan datasets, we follow the preprocessing pipeline outlined in Liu et al.\cite{liu2023common}. The REST-meta-MDD dataset is preprocessed as described in Yan et al.\cite{yan2019reduced}. The SMHC, HBN, and DAMOMRI datasets are preprocessed using the Gretna toolbox~\cite{wang2015gretna}, which includes discarding the first 10 volumes, slice timing correction, motion correction, covariate removal, and temporal filtering.

\subsection{Experimental Setup}


\paragraph{Data Split}
The 17 datasets are divided into two groups for pretraining and downstream evaluation. We choose 6 datasets for evaluation with various tasks including neurological disorder identification (AD, ASD, ADHD), sex prediction and age prediction. Specifically, ADNI, ABIDE and ADHD are used for internal evaluation. For each of these datasets, 80\% of the subjects are used during the pretraining stage and subsequently fine-tuned for downstream tasks, while the remaining subjects are split into validation and test sets with a ratio of 10\%/10\%. We repeat the experiments 10 times with different random seeds and report the average performance. In addition, HCP-Gender, HCP-Age, and Huashan are used for external evaluation. Each dataset is independently split into training, validation, and test sets with an 8:1:1 ratio. For these external datasets, models are finetuned and evaluated using 10-fold cross-validation with a fixed random seed, and the average results across folds are reported.

\begin{table*}[t]
\caption{Results (average AUC $\pm$ standard deviation) on 6 datasets by averaging all available atlases. The best result is highlighted in \textbf{bold} while the runner-up is highlighted in \underline{underlined}.}
\centering
\setlength\tabcolsep{4.5pt}
\begin{tabular}{lccccccccc}
\hline
\multirow{2}{*}{Method} & \multirow{2}{*}{Pretrained} & \multicolumn{4}{c}{Internal Tests}                        & \multicolumn{4}{c}{External Tests}                        \\ \cmidrule(lr){3-6} \cmidrule(lr){7-10}
                        &                             & ADHD         &
                        ADNI         & ABIDE        &  Avg.  & HCP-Gender    & HCP-Age      & Huashan      & Avg.  \\ \hline
GCN~\cite{kipf2017semi}                     &                    \XSolidBrush         & 54.66 $\pm$ 6.38 & 66.06 $\pm$ 2.20 & 62.58 $\pm$ 4.95 &  61.1  & 74.13 $\pm$ 4.99  & 54.43 $\pm$ 8.25 & 54.73 $\pm$ 3.20 & 61.1  \\
BrainNetCNN~\cite{kawahara2017brainnetcnn}             &                    \XSolidBrush         & 55.10 $\pm$ 5.32 & \underline{68.91} $\pm$ 2.06 & 67.18 $\pm$ 7.67 &  63.73 & 73.35 $\pm$ 14.08 & 54.81 $\pm$ 8.24 & 59.20 $\pm$ 4.73 & 62.46 \\
BrainGNN~\cite{li2021braingnn}                &           \XSolidBrush                  & 55.86 $\pm$ 5.74 & 63.23 $\pm$ 3.10 & 60.41 $\pm$ 4.02 &  59.83 & 72.17 $\pm$ 10.30 & 52.98 $\pm$ 5.41 & 55.82 $\pm$ 4.85 & 60.32 \\
BNT~\cite{kan2022brain}                     &             \XSolidBrush                & 53.08 $\pm$ 7.21 & 61.04 $\pm$ 6.78 & 64.46 $\pm$ 7.22 &  59.53 & 76.57 $\pm$ 11.21 & 51.26 $\pm$ 5.98 & \underline{59.91} $\pm$ 3.68 & 62.58 \\
ContrastPool~\cite{xu2024contrastive}            &              \XSolidBrush               & 59.65 $\pm$ 5.19 & 64.65 $\pm$ 3.66 & 68.69 $\pm$ 2.72 &  64.33 & 80.00 $\pm$ 5.80  & \underline{56.30} $\pm$ 5.58 & 58.35 $\pm$ 6.11 & 64.88 \\
BQN~\cite{yangwe}                     &           \XSolidBrush                  & \underline{60.95} $\pm$ 3.09 & 66.09 $\pm$ 2.90 & 68.75 $\pm$ 5.04 &  \underline{65.26} & 82.60 $\pm$ 5.44  & 53.18 $\pm$ 7.46 & 58.98 $\pm$ 4.85 & 64.92 \\ 
BrainLM~\cite{caro2024brainlm}                 &                \Checkmark             & 59.35 $\pm$ 6.53 & 55.51 $\pm$ 5.68 & 61.39 $\pm$ 4.88 &  58.75 & 70.95 $\pm$ 4.20  & 51.97 $\pm$ 6.68 & 52.91 $\pm$ 5.34 & 58.61 \\
BrainJEPA~\cite{dong2024brain}               &                \Checkmark             & 57.75 $\pm$ 4.80 & 57.28 $\pm$ 5.89 & 61.71 $\pm$ 7.14 &  58.91 & 73.01 $\pm$ 4.72  & 54.79 $\pm$ 6.98 & 54.76 $\pm$ 5.58 & 60.85 \\
PTGB~\cite{yang2023ptgb}                    &             \Checkmark                & 58.55 $\pm$ 3.91 & 58.15 $\pm$ 7.24 & 65.00 $\pm$ 6.62 &  60.57 & 75.12 $\pm$ 5.35  & 52.29 $\pm$ 6.04 & 52.19 $\pm$ 5.18 & 59.87 \\
BrainMass~\cite{yang2024brainmass}               &              \Checkmark               & 53.92 $\pm$ 7.24 & 55.23 $\pm$ 5.86 & 64.53 $\pm$ 6.09 &  57.89 & 74.01 $\pm$ 5.43  & 52.63 $\pm$ 6.54 & 52.11 $\pm$ 4.90 & 59.59 \\
BrainGFM~\cite{wei2026a}                &           \Checkmark                  & 53.79 $\pm$ 7.33 & 67.82 $\pm$ 2.85 & \underline{69.52} $\pm$ 2.57 &  63.71 & \underline{84.61} $\pm$ 3.70  & 55.05 $\pm$ 6.59 & 57.69 $\pm$ 4.90 & \underline{65.78} \\ \hline
MV-BrainFM (Ours)                 &               \Checkmark              & \textbf{61.18} $\pm$ 5.65 & \textbf{70.02} $\pm$ 2.27$^{*}$ & \textbf{74.98} $\pm$ 5.19$^{*}$ &  \textbf{68.73} & \textbf{86.93} $\pm$ 3.22$^{*}$  & \textbf{59.77} $\pm$ 5.75$^{*}$ & \textbf{61.64} $\pm$ 5.37 & \textbf{69.44} \\ \hline
\end{tabular}
\begin{flushleft}
$^{*}$ indicates statistically significant improvement over the best competing method ($p<0.05$, paired t-test).
\end{flushleft}
\label{tab:main_results}
\vspace{-0.2cm}
\end{table*}

\paragraph{Baselines}
We select various baselines, including (1) \textbf{general-purpose GNN}: GCN~\cite{kipf2017semi}; (2) \textbf{end-to-end models tailored brain networks}: BrainNetCNN~\cite{kawahara2017brainnetcnn}, BrainGNN~\cite{li2021braingnn}, BNT~\cite{kan2022brain}, ContrastPool~\cite{xu2024contrastive}, BQN~\cite{yangwe}; (3) \textbf{multi-atlas methods}: MGRL~\cite{cooper2017reduced}, METAFormer~\cite{mahler2023pretraining}, AIDFusion~\cite{xu2025multi}; (4) \textbf{time-series-based brain foundation models}: BrainLM~\cite{caro2024brainlm}, Brain-JEPA~\cite{dong2024brain}; and (5) \textbf{graph-based brain network foundation models}: PTGB~\cite{yang2023ptgb}, BrainMass~\cite{yang2024brainmass}, BrainGFM~\cite{wei2026a}. 

Existing brain foundation models differ in how they handle heterogeneity across fMRI datasets. Time-series-based models are pre-trained directly on fMRI time-series with Transformer-style architectures, and they generally require preprocessing into fixed-length temporal windows and a unified atlas-specific input format. By contrast, graph-based methods operate on functional connectivity or brain graphs, making them more easily handle variable-length fMRI data after graph construction: PTGB trains a linear encoder for each dataset to address cross-atlas heterogeneity; BrainMass proposes different dropping percentages to account for time-series length variability, and employs a comprehensive pre-training pipeline to learn perturbation-robust representations, enabling frozen-encoder finetuning; and BrainGFM introduces graph prompts to align multi-atlas datasets within a unified modeling framework. For fair comparison, we use single-atlas datasets for the main experiments, although MV-BrainFM is able to handle multi-atlas input.

\paragraph{Hyperparameter Selection}
We pretrain and fine-tune the network end-to-end using the Adam optimizer~\cite{kingma2014adam}. 
The learning rate for pretraining is fixed at $1\times10^{-4}$, while the initial learning rate for fine-tuning is selected from $\{1\times10^{-3}, 5\times10^{-4}, 2\times10^{-4}, 1\times10^{-4}, 5\times10^{-5}\}$ based on validation performance. 
To improve training stability and convergence, we adopt a warmup–cosine learning rate schedule for finetuning stage. 

The model is pretrained for 20 epochs and subsequently fine-tuned on downstream tasks for an additional 50 epochs. 
Unless otherwise specified, the backbone encoder consists of $L=4$ Transformer layers, each with 8-head DASA. 
The number of learnable frequency vectors $K$ in the atlas encoder (Eq. (\ref{eq:freq_vec})) is set to 64.
The number of supernodes $N_{super}$ in Eq. (~\ref{eq:supernode}) is set to 50 while the number of prototypes $P$ used in the clustering consistency in Eq.~(\ref{eq:prototype}) is set to 16. During pretraining, a dropout rate of 0.5 is applied to mitigate overfitting, while dropout is disabled during fine-tuning. 
All experiments are conducted on a Linux server equipped with an AMD EPYC 7402 24-core CPU and an NVIDIA A100 GPU.

\begin{figure*}[t]
\centering
\includegraphics[width=\textwidth]{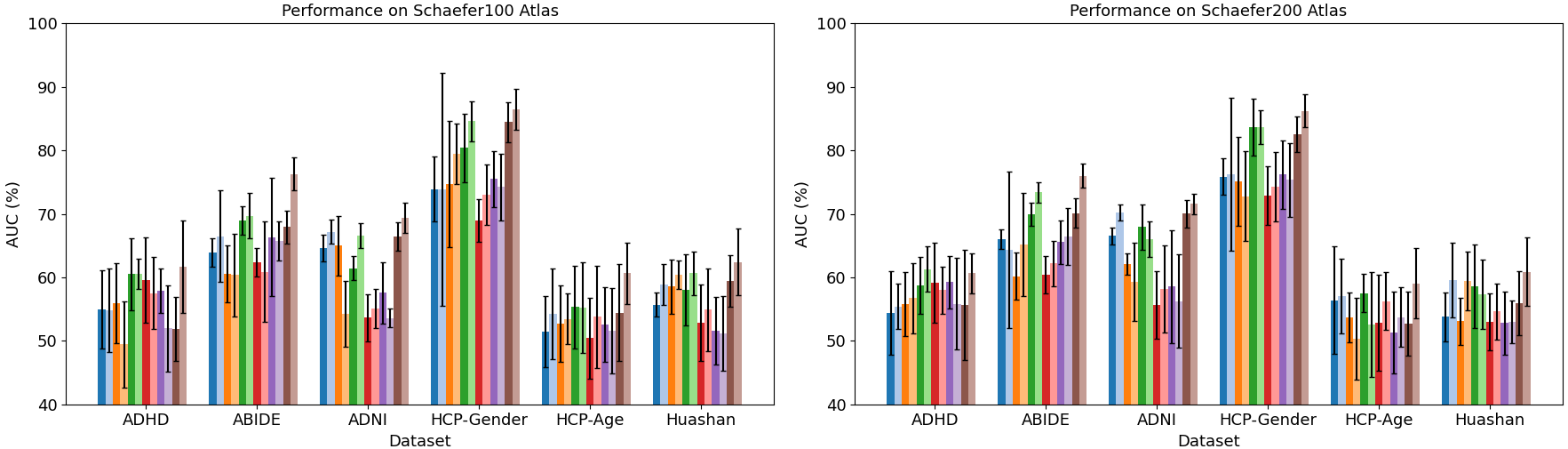}
\includegraphics[width=\textwidth]{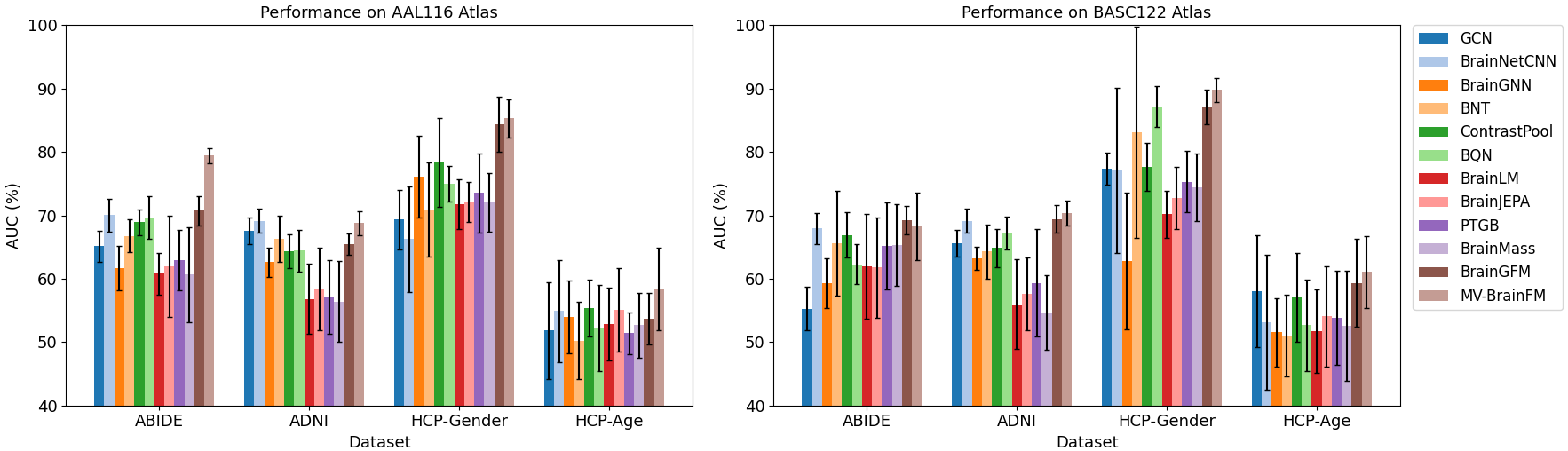}
\caption{Results on all 4 atlases. Note that BASC122 is the external atlas, which the models have never seen during pretraining.}
\label{fig:all_atlases}
\end{figure*}

\subsection{Single Atlas Results}

We evaluate the pretrained MV-BrainFM on 6 downstream datasets constructed using 4 brain atlases. Three of the atlases appear during the pretraining stage, while the remaining atlas (BASC122) is intentionally excluded from pretraining to assess the cross-atlas transfer capability of the proposed framework. For fair comparison with baseline methods, each dataset is evaluated independently under each atlas rather than using multi-view inputs during inference. This experimental setup enables us to assess both the effectiveness of the pretrained representations and the ability of the model to generalize to previously unseen atlas configurations.

Table~\ref{tab:main_results} summarizes the classification performance of all methods, reported as the average AUC across all available atlases for each dataset. Overall, MV-BrainFM achieves the best performance on all six datasets and obtains the highest average AUC in both the internal and external evaluations. Specifically, MV-BrainFM improves the internal average AUC from the previous best result of 65.26\% (BQN) to 68.73\%, and improves the external average AUC from 65.78\% (BrainGFM) to 69.44\%. Moreover, MV-BrainFM achieves statistically significant improvements over the strongest competing methods on several datasets, including ADNI, ABIDE, HCP-Gender, and HCP-Age ($p<0.05$). These results demonstrate the effectiveness of the proposed multi-view pretraining strategy in learning robust brain network representations.

Compared with conventional graph learning models such as GCN, BrainNetCNN, BrainGNN, and BNT, MV-BrainFM consistently achieves superior performance across all datasets. These models are typically trained from scratch on a single dataset and rely on one specific atlas representation, which limits their ability to exploit complementary information across different brain parcellations. In contrast, MV-BrainFM leverages multi-atlas pretraining and learns atlas-invariant representations that transfer effectively to downstream tasks.

We further compare MV-BrainFM with several recently proposed brain foundation models, including BrainLM, BrainJEPA, PTGB, BrainMass, and BrainGFM. Although these models are pretrained on large-scale neuroimaging data, their downstream performance is often comparable to or even worse than models specifically designed for individual tasks. This observation suggests that existing brain foundation models may not fully capture the structural characteristics of brain networks required for disease prediction tasks. In contrast, MV-BrainFM explicitly models brain networks as graphs and incorporates anatomical priors through distance-aware attention, allowing it to better exploit the topology of functional connectivity patterns during pretraining.

To further examine the cross-atlas robustness of different methods, Fig.~\ref{fig:all_atlases} reports the performance of all models on each individual atlas. As shown in the figure, MV-BrainFM consistently achieves the best or near-best performance across all four atlases (Schaefer100, Schaefer200, AAL116, and BASC122), despite their different spatial resolutions and ROI definitions. Importantly, the performance gain remains evident even on the atlas that was not used during pretraining, demonstrating the strong cross-atlas transfer capability of the proposed framework.

Another interesting observation from the detailed results is that no single atlas consistently yields the best performance across all datasets. Instead, different atlases perform better for different tasks and datasets. This observation aligns with previous findings in neuroscience, where the choice of brain parcellation has been shown to influence downstream analyses and different atlases may capture complementary aspects of brain organization~\cite{arslan2018human}. Therefore, relying on a single atlas may limit model performance, while integrating multiple atlases can provide richer and more robust representations of brain connectivity.

\begin{figure*}[t]
\centering
\includegraphics[width=.85\textwidth]{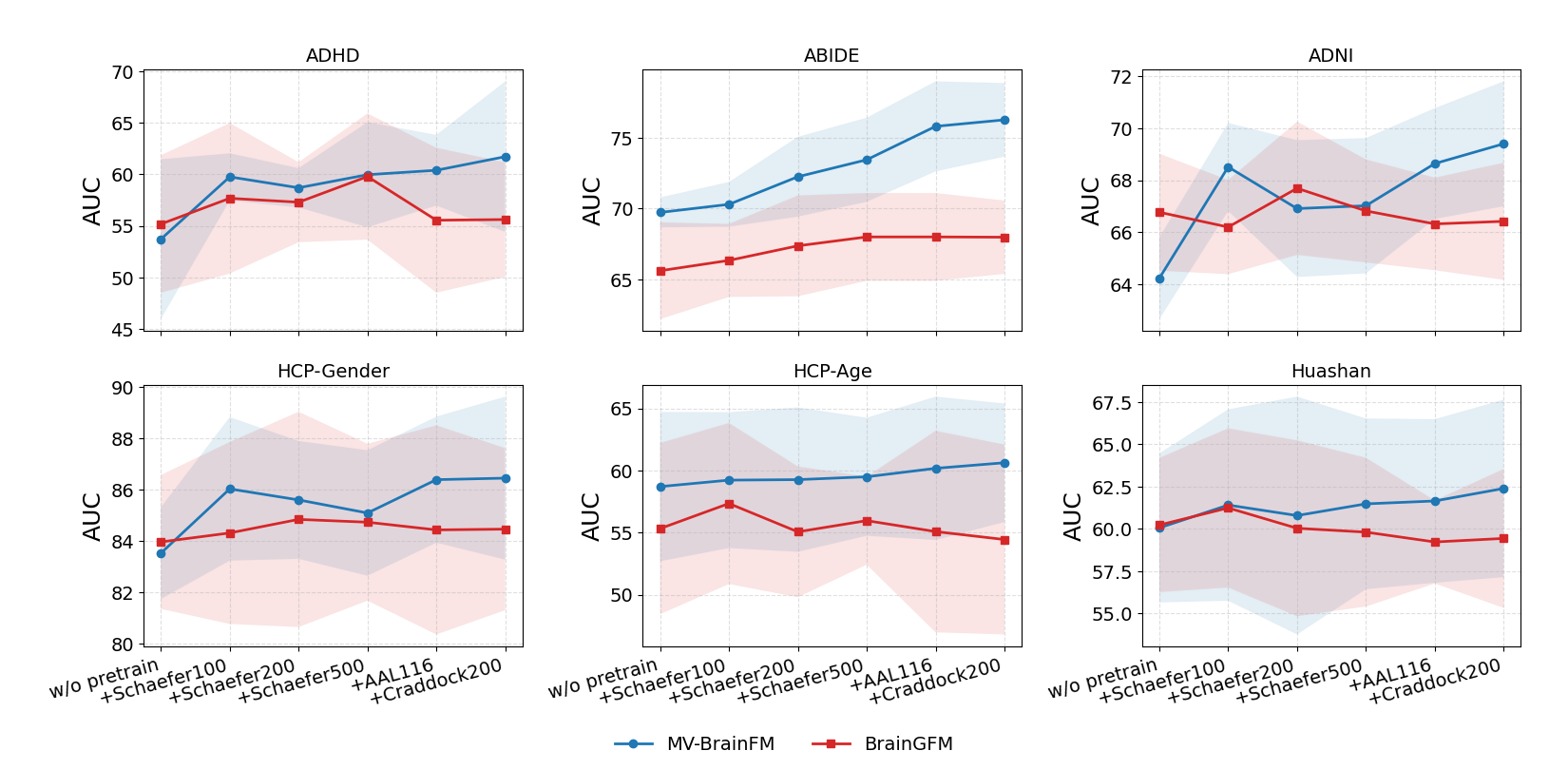}
\caption{Scaling behavior with increasing pretraining atlases. We progressively increase the number of atlases used during pretraining and evaluate the downstream performance on 6 tasks using the Schaefer100 atlas. The results show that MV-BrainFM consistently benefits from larger and more diverse pretraining data, demonstrating clearer scaling behavior than BrainGFM.}
\label{fig:scale}
\end{figure*}

\subsection{Multi-atlas Results}

We further evaluate the ability of MV-BrainFM to directly leverage multiple atlases as input. Specifically, we adopt a simple yet effective strategy by summing the embeddings from different atlases and applying a linear prediction head for downstream tasks. As shown in Table~\ref{tab:multi_atlas}, MV-BrainFM consistently outperforms all multi-atlas baselines across all four datasets when jointly using four atlases (Schaefer100, Schaefer200, AAL116, and BASC122). Notably, the multi-atlas performance of MV-BrainFM surpasses its corresponding single-atlas results on all datasets, demonstrating the benefit of integrating complementary information from multiple brain parcellations. In contrast to existing methods that require specialized fusion designs, MV-BrainFM enables straightforward and effective multi-atlas integration due to its unified representation space learned during pretraining. These results demonstrate that MV-BrainFM provides a flexible and scalable framework that can seamlessly incorporate an arbitrary number of atlases, making it particularly suitable for large-scale and heterogeneous brain network analysis.

\begin{table}[t]
\renewcommand{\arraystretch}{1.4}
\caption{Multi-atlas AUC performance comparison using four atlases (Schaefer100, Schaefer200, AAL116, and BASC122). MV-BrainFM achieves superior performance over existing multi-atlas baselines on all datasets.}
\scalebox{0.85}{
\begin{tabular}{lcccc}
\hline
\multirow{2}{*}{Method} & \multicolumn{2}{c}{Internal Tests}                        & \multicolumn{2}{c}{External Tests}                        \\ \cmidrule(lr){2-3} \cmidrule(lr){4-5}
    & ABIDE        & ADNI         & HCP-Gender   & HCP-Age      \\ \hline
MGRL~\cite{cooper2017reduced}       & 66.99 $\pm$ 4.86 & 59.57 $\pm$ 1.75 & 76.01 $\pm$ 8.40 & 56.78 $\pm$ 6.17 \\
METAFormer~\cite{mahler2023pretraining} & 75.87 $\pm$ 3.39 & 66.40 $\pm$ 1.24 & 84.17 $\pm$ 4.39 & 61.40 $\pm$ 4.69 \\
AIDFusion~\cite{xu2025multi}  & 75.19 $\pm$ 1.38 & 66.59 $\pm$ 2.43 & 85.34 $\pm$ 5.78 & 59.77 $\pm$ 5.35 \\
MV-BrainFM & \textbf{77.93} $\pm$ 2.36 & \textbf{72.63} $\pm$ 0.90 & \textbf{86.64} $\pm$ 4.54 & \textbf{61.47} $\pm$ 5.47 \\ \hline
\end{tabular}
}
\label{tab:multi_atlas}
\end{table}

\begin{figure*}[t]
\centering
\includegraphics[width=.85\textwidth]{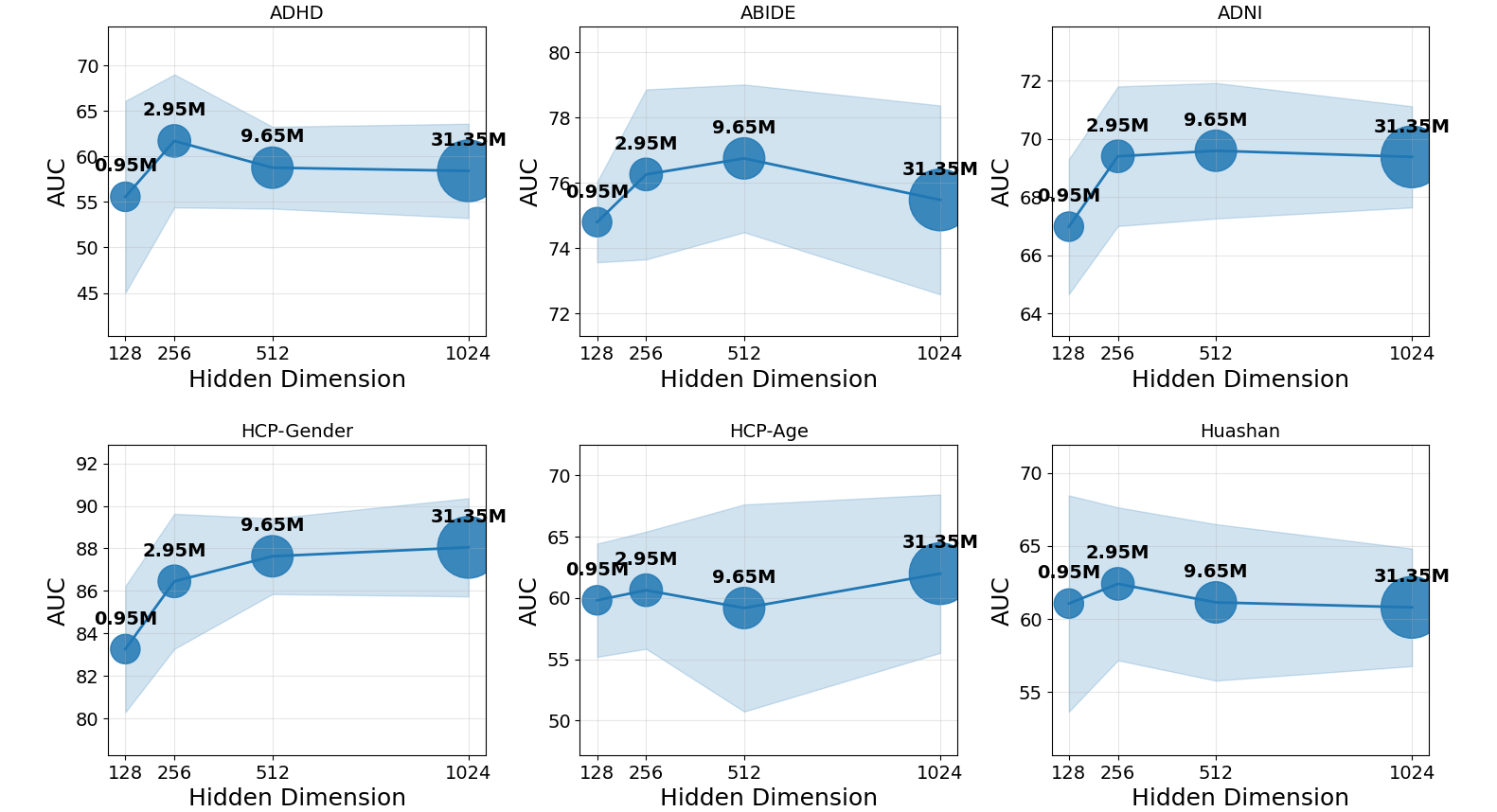}
\caption{AUC performance of MV-BrainFM using the Schaefer100 atlas under different hidden dimensions (0.95M–31.35M parameters). Increasing model capacity improves performance from very small models, but larger models do not consistently yield additional gains, indicating that data scale may be a more critical factor than parameter scale.}
\label{fig:scale_param}
\end{figure*}

\subsection{Scaling Study}

We further investigate how the performance of brain network foundation models changes as the amount of pretraining data increases. Specifically, we progressively expand the set of atlases used during pretraining and evaluate downstream performance on 6 tasks using the Schaefer100 atlas. The atlases are introduced sequentially in the order of Schaefer100, Schaefer200, Schaefer500, AAL116, and Craddock200, allowing us to analyze how model performance scales with increasing diversity of pretraining data.

As shown in Fig.~\ref{fig:scale}, both models benefit from pretraining compared with directly fine-tuning on downstream tasks without pretraining, demonstrating the effectiveness of representation learning on large-scale brain network data. Note that MV-BrainFM exhibits a more stable and consistent scaling trend as additional atlases are incorporated, while BrainGFM shows more fluctuating behavior.

More specifically, MV-BrainFM tends to achieve larger improvements on the internal datasets (ADHD, ABIDE, and ADNI) than on the external datasets (HCP-Gender, HCP-Age, and Huashan). This observation suggests that incorporating multiple views of the same subject during pretraining helps the model learn more discriminative connectivity patterns rather than overfitting to limited data representations. By enforcing cross-view consistency, MV-BrainFM effectively leverages complementary information across different brain parcellations while maintaining stable representations.

In contrast, BrainGFM occasionally experiences performance degradation when additional atlases are introduced. Since BrainGFM treats different atlases independently and does not explicitly enforce multi-view consistency, the model may struggle to reconcile heterogeneous graph structures during pretraining. Consequently, introducing networks constructed from additional atlases may increase representation variability instead of improving representation quality.

We also observe a slight performance drop for MV-BrainFM when Schaefer200 is added after Schaefer100 for several datasets. One possible explanation is that the Schaefer200 atlas introduces more than 12,000 additional subjects from the HBN dataset that are not included in the Schaefer100 pretraining set. This substantial increase of data from a single atlas may temporarily bias the pretraining process toward the Schaefer200 representation. Interestingly, this effect is alleviated once additional atlases such as AAL116 and Craddock200 are incorporated, suggesting that increasing atlas diversity helps rebalance the multi-view learning process and restore performance gains.

We further analyze the impact of model capacity in Fig.~\ref{fig:scale_param}, where the hidden dimension varies from 128 to 1024, corresponding to model sizes ranging from 0.95M to 31.35M parameters. Increasing the hidden dimension from 128 to 256 generally improves performance across most datasets, indicating that extremely small models may lack sufficient representation capacity. However, further increasing the model size beyond moderate scales (e.g., 512 or 1024) does not consistently improve performance and may even slightly degrade results on some datasets. This observation suggests that the current performance bottleneck lies more in the scale and diversity of training data rather than model capacity. Although our study already leverages the largest brain network dataset reported in the existing literature, the results indicate that the proposed framework could still benefit from training on even larger datasets. Moreover, the observation that larger models do not necessarily lead to better performance is consistent with findings reported in other neuroimaging foundation models~\cite{jiang2024large,banville2025scaling}. Considering both efficiency and effectiveness, we adopt the configuration with 2.95M parameters as the default setting in MV-BrainFM.

Overall, these results demonstrate that MV-BrainFM scales more effectively with increasing pretraining data and can better exploit multi-atlas information. The proposed cross-view consistency learning strategy enables the model to integrate heterogeneous brain network views more robustly, leading to improved scalability and generalization compared with existing brain network foundation models.

\begin{figure*}[t]
\centering
\includegraphics[width=.9\textwidth]{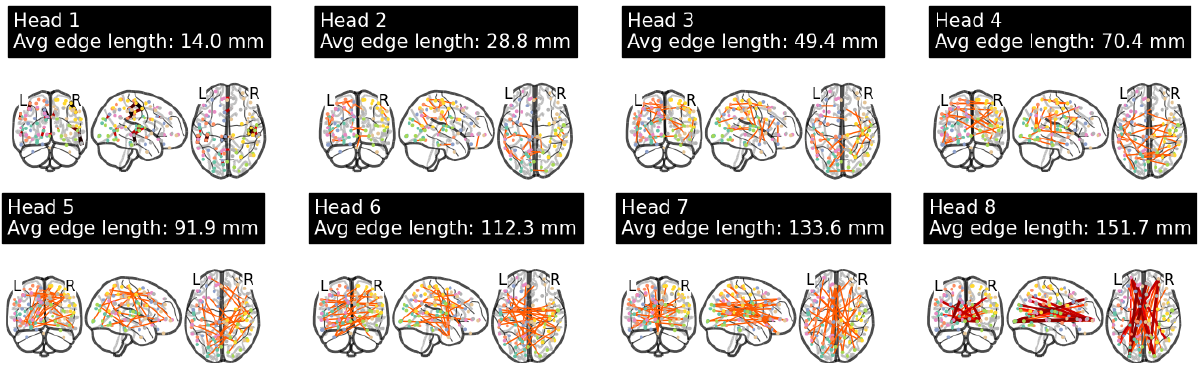}
\caption{Visualization of the top 20 edges with the largest distance bias for each attention head. Different heads consistently focus on connections at distinct spatial scales.}
\label{fig:vis_bias}
\vspace{-0.5cm}
\end{figure*}

\begin{figure*}[t]
\centering
\subfloat[ADHD]{
    \includegraphics[width=0.234\textwidth]{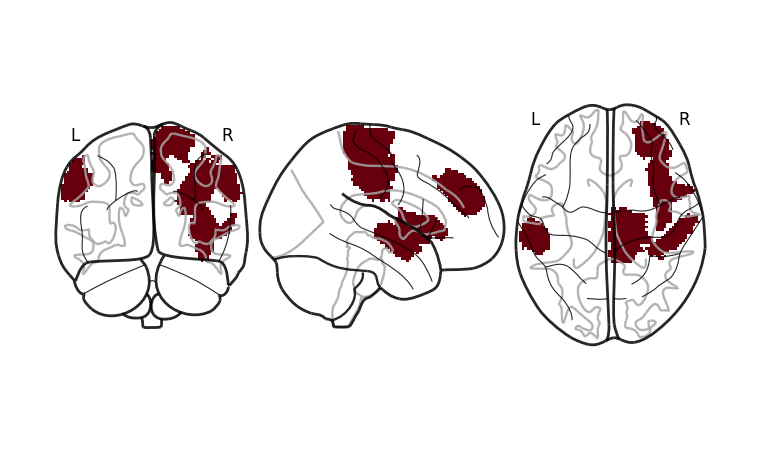}
}
\hfill
\subfloat[ABIDE]{
    \includegraphics[width=0.234\textwidth]{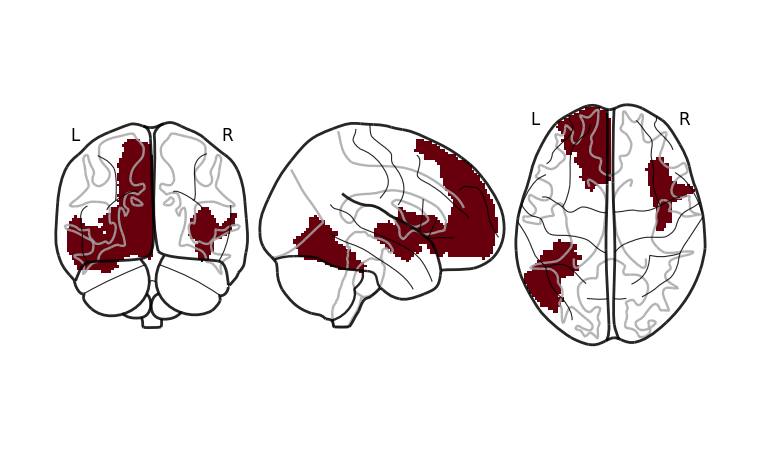}
}
\hfill
\subfloat[ADNI]{
    \includegraphics[width=0.234\textwidth]{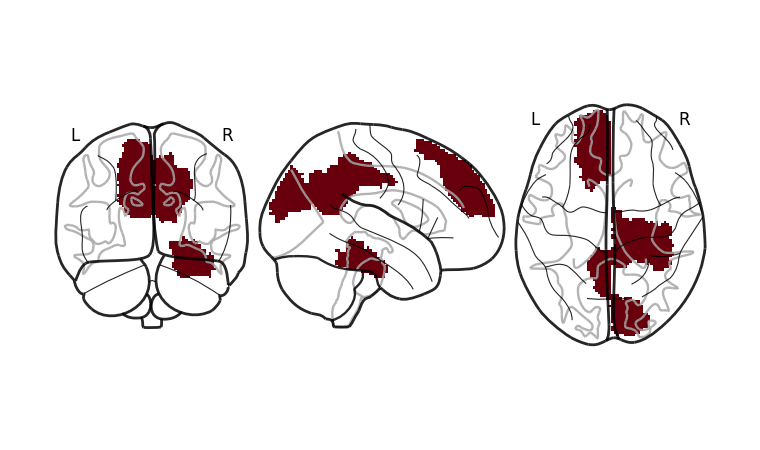}
}
\hfill
\subfloat[Huashan]{
    \includegraphics[width=0.234\textwidth]{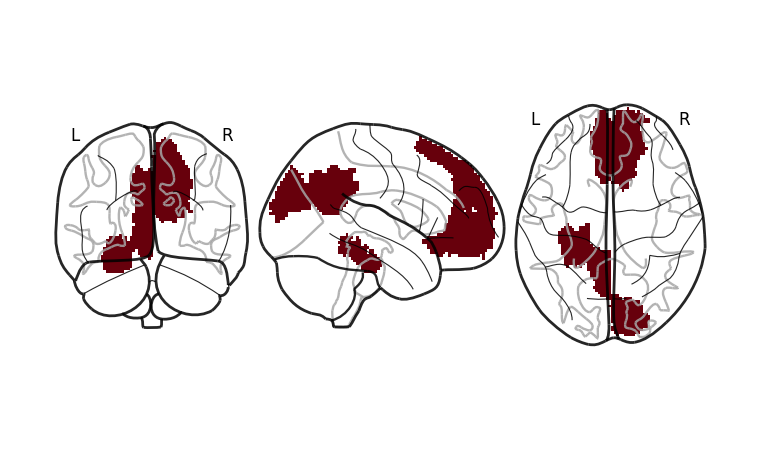}
}
\caption{Visualization of salient brain regions identified by MV-BrainFM across different neurological disorders. The highlighted regions correspond to ROIs with the highest attention weights and represent potential disease-related biomarkers for ADHD, ASD (ABIDE), AD/MCI (ADNI and Huashan).}
\label{fig:biomarker_vis}
\end{figure*}

\begin{table*}[t]
\renewcommand{\arraystretch}{1.2}
\caption{Time cost (seconds) and number of parameters for brain network foundation models on datasets with Schaefer100 atlas. The finetune time reports the 50 epochs running time for a single fold including training, validation, and test.}
\centering
\begin{tabular}{lccccccccc}
\hline
\multirow{2}{*}{Models} & \multirow{2}{*}{\#Params} & \multicolumn{2}{c}{Pretrain} & \multicolumn{6}{c}{Finetune Time (s)}                      \\ \cmidrule(lr){3-4} \cmidrule(lr){5-10}
                  &                           & \#Epoch        & Time (s)        & ADHD  & ABIDE & ADNI  & HCP-Gender & HCP-Age & Huashan \\ \hline
BrainLM           & 12.78M                          & 50             & 3,989       & 430   & 609   & 785   & 632        & 605     & 462     \\
Brain-JEPA        & 21.69M                          & 50             & 4,124       & 1,038 & 1,573 & 1,814 & 1,630      & 1,275   & 1,126   \\
PTGB              & 0.55M                          & 70            & 1,574       & 23    & 15    & 22    & 26         & 30      & 29      \\
BrainMass         & 29.35M                          & 100            & 6,988       & 74    & 80    & 102   & 102        & 117     & 84      \\
BrainGFM          & 10.31M                    & 20             & 3,709        & 46    & 58    & 83    & 63         & 63      & 47      \\
MV-BrainFM (ours)            & 2.95M                     & 20             & 935         & 13    & 18    & 24    & 30         & 30      & 13      \\ \hline
\end{tabular}
\label{tab:time}
\end{table*}

\subsection{Model Interpretation}

\paragraph{Distance Bias}
To better understand how the proposed distance-aware attention mechanism influences representation learning, we visualize the top 20 edges with the largest distance bias for each attention head, as shown in Fig.~\ref{fig:vis_bias}. Interestingly, different heads consistently focus on connections at distinct spatial scales. For example, early heads mainly capture short-range interactions with average edge lengths around $14$--$30$\,mm, which typically correspond to local intra-network connections within nearby cortical regions. As the head index increases, the preferred interaction distance gradually expands, with later heads emphasizing medium- and long-range connections exceeding $100$\,mm, which often link distant functional systems across hemispheres.

This observation suggests that the proposed Gaussian distance bias successfully encourages different attention heads to specialize in modeling connectivity patterns at multiple anatomical scales. Such a multi-scale decomposition aligns well with known properties of brain organization, where local circuits support specialized processing while long-range connections integrate distributed functional networks~\cite{bullmore2012economy}. Importantly, this behavior emerges automatically from the learned distance-aware bias without requiring explicit supervision, demonstrating that MV-BrainFM can leverage anatomical priors to discover biologically meaningful connectivity patterns.

\paragraph{Biomarker Detection}
Identifying salient brain regions associated with model predictions is essential for discovering potential neural biomarkers. In this study, we leverage the built-in interpretability of MV-BrainFM to explore disease-specific biomarker patterns. Specifically, we visualize the most influential ROIs for AD, ASD, ADHD, MCI using the Nilearn toolbox~\cite{abraham2014machine}. For each prediction, we select the ROIs with the top 5 attention weights.

As shown in Fig.~\ref{fig:biomarker_vis}, several salient regions emerge that are consistent with previously reported disease-related brain alterations. For ADHD, MV-BrainFM identifies several regions within the dorsal attention network, including the right precentral gyrus and the bilateral posterior parietal cortex. These regions are known to play important roles in attentional control and executive function, and have been repeatedly implicated in ADHD-related neural alterations~\cite{ma2016ventral,o2018neural}.

For the ABIDE dataset (ASD), MV-BrainFM highlights regions including the left prefrontal cortex, the left somatomotor cortex, and the right precentral cortex. These regions have been previously implicated in autism studies~\cite{assaf2010abnormal,wang2006neural,soulieres2009enhanced}, where abnormalities in social cognition, sensory processing, and attention regulation have been reported.

For AD and MCI, MV-BrainFM highlights regions such as the right visual cortex and the right frontal eye field within the dorsal attention network in the ADNI dataset, as well as the left parietal cortex and the right visual cortex in the Huashan dataset. These findings are consistent with prior studies showing that Alzheimer’s disease is strongly associated with disruptions in visual processing pathways and the Default Mode Network~\cite{damoiseaux2012functional,bosch2010cognitive,deng2016mapping}.

Overall, these results demonstrate that MV-BrainFM can identify clinically meaningful brain regions associated with neurological disorders, suggesting its potential utility for biomarker discovery and neurobiological interpretation.

\subsection{Efficiency Analysis}

Table~\ref{tab:time} reports the number of parameters, pretraining cost, and finetuning time of different brain network foundation models. The finetuning time corresponds to the running time for a single cross-validation fold, including training, validation, and testing.

Overall, MV-BrainFM demonstrates significantly higher computational efficiency compared with existing brain foundation models. In terms of model size, MV-BrainFM contains only 2.95M parameters, which is substantially smaller than most of the other large-scale architectures. Despite its compact design, MV-BrainFM achieves superior predictive performance, indicating that the proposed architecture provides a favorable balance between model capacity and computational efficiency.

\begin{table*}[t]
\centering
\caption{Ablation study on  components in MV-BrainFM on datasets with the Schaefer100 atlas.}
\begin{tabular}{ccccccccccc}
\hline
Atlas Encoder & Distance Bias & $\mathcal{L}_{\text{rec}}$ & $\mathcal{L}_{\text{cc}}$ & $\mathcal{L}_{\text{ent}}$ & ADHD         & ABIDE        & ADNI         & HCP-Gender   & HCP-Age      & Huashan      \\ \hline
\Checkmark             & \Checkmark             &        &       &        & 54.70 $\pm$ 7.76 & 69.72 $\pm$ 1.06 & 64.23 $\pm$ 1.57 & 83.52 $\pm$ 1.79 & 58.70 $\pm$ 6.01 & 60.06 $\pm$ 4.40 \\
              &               & \Checkmark      & \Checkmark     & \Checkmark      &           57.26 $\pm$ 6.47   &   73.38 $\pm$ 1.68           &  67.74 $\pm$ 1.34            &  85.54 $\pm$ 2.52            &      57.99 $\pm$ 5.88        &      60.71 $\pm$ 6.23        \\
              & \Checkmark             & \Checkmark      & \Checkmark     & \Checkmark      & 58.10 $\pm$ 4.97 & 74.38 $\pm$ 2.49 & 68.20 $\pm$ 2.10 & 85.36 $\pm$ 2.67 & 58.72 $\pm$ 6.22 & 61.61 $\pm$ 4.48 \\
\Checkmark             &               & \Checkmark      & \Checkmark     & \Checkmark      &      58.99 $\pm$ 6.45        &     75.27 $\pm$ 2.49         &      69.24 $\pm$ 1.63        &         86.77 $\pm$ 1.77      &       59.67 $\pm$  5.42       &     61.31 $\pm$ 5.08       \\
\Checkmark             & \Checkmark             &        & \Checkmark     & \Checkmark      & 61.18 $\pm$ 3.54 & 72.79 $\pm$ 1.48 & 66.19 $\pm$ 2.79 & 84.41 $\pm$ 2.05 & 59.11 $\pm$ 4.21 & 61.64 $\pm$ 6.43 \\
\Checkmark             & \Checkmark             & \Checkmark      &       & \Checkmark      & 59.76 $\pm$ 1.85 & 73.22 $\pm$ 1.96 & 69.30 $\pm$ 0.89 & 86.38 $\pm$ 2.90 & 59.01 $\pm$ 5.15 & 61.83 $\pm$ 5.22 \\
\Checkmark             & \Checkmark             & \Checkmark      & \Checkmark     &        & 57.53 $\pm$ 7.86 & 73.20 $\pm$ 1.62 & 64.54 $\pm$ 1.55 & 83.93 $\pm$ 1.31 & 59.75 $\pm$ 7.89 & 60.94 $\pm$ 3.60 \\
\Checkmark             & \Checkmark             & \Checkmark      & \Checkmark     & \Checkmark      & \textbf{61.71} $\pm$ 7.31 & \textbf{76.26} $\pm$ 2.60 & \textbf{69.40} $\pm$ 2.40 & \textbf{86.45} $\pm$ 3.18 & \textbf{60.62} $\pm$ 4.77 & \textbf{62.39} $\pm$ 5.24 \\ \hline
\end{tabular}
\label{tab:ablation}
\end{table*}

Regarding pretraining cost, MV-BrainFM also shows a clear advantage. The total pretraining time of MV-BrainFM is 935 seconds for 20 epochs, which is substantially lower than most competing foundation models. This efficiency arises from two aspects. First, the proposed architecture adopts a lightweight Transformer design with distance-aware attention, resulting in a relatively small parameter size. Second, unlike many existing brain foundation models that pretrain sequentially on datasets constructed with different atlases, MV-BrainFM adopts a unified multi-view training paradigm. In this paradigm, mini-batches from all datasets are shuffled together and multiple atlas views of the same subject are optimized simultaneously. As a result, the model can learn from all datasets within a single training process rather than repeatedly retraining on each dataset, which significantly reduces the total pretraining time.

The efficiency advantage is even more pronounced during finetuning. Across all 6 downstream datasets, MV-BrainFM consistently requires the shortest training time. For example, the finetuning time on the ADNI dataset is only 24 seconds, compared with 83 seconds for BrainGFM, 102 seconds for BrainMass, and several thousand seconds for Brain-JEPA. Similar improvements are observed across the remaining datasets, where MV-BrainFM reduces the finetuning time by several times compared with existing foundation models.

These results demonstrate that MV-BrainFM not only improves predictive performance but also significantly reduces computational cost. Such efficiency is particularly important for large-scale neuroimaging studies where models must be trained and evaluated across multiple datasets and atlas configurations.

\begin{figure*}[h]
\centering
\subfloat[{\scriptsize The sensitivity analysis for the number of heads in MV-BrainFM.}]
{
    \includegraphics[width=0.43\textwidth]{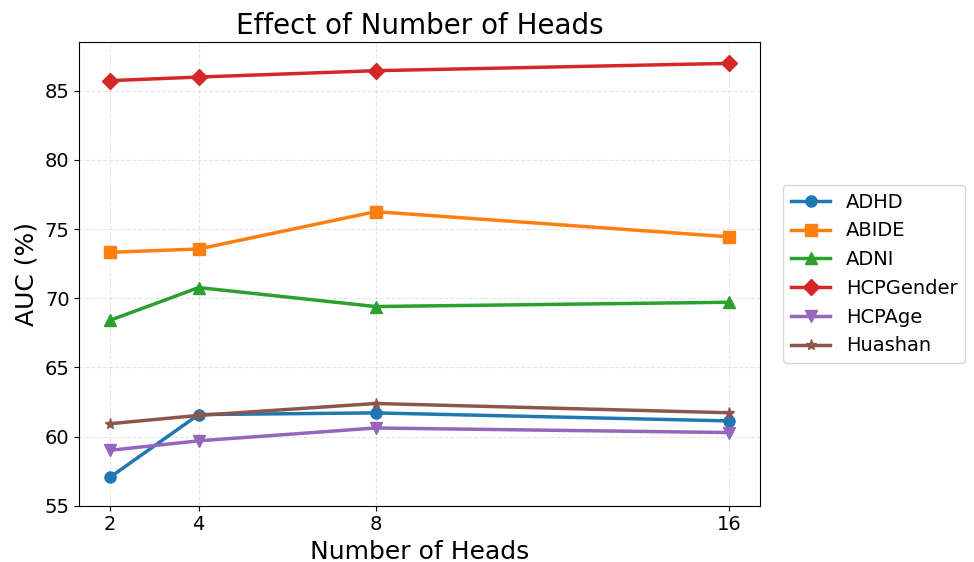}
    \label{fig:hyperparam_nhead}
}
\hfill
\subfloat[{\scriptsize The sensitivity analysis for the number of prototypes in MV-BrainFM.}]
{
    \includegraphics[width=0.43\textwidth]{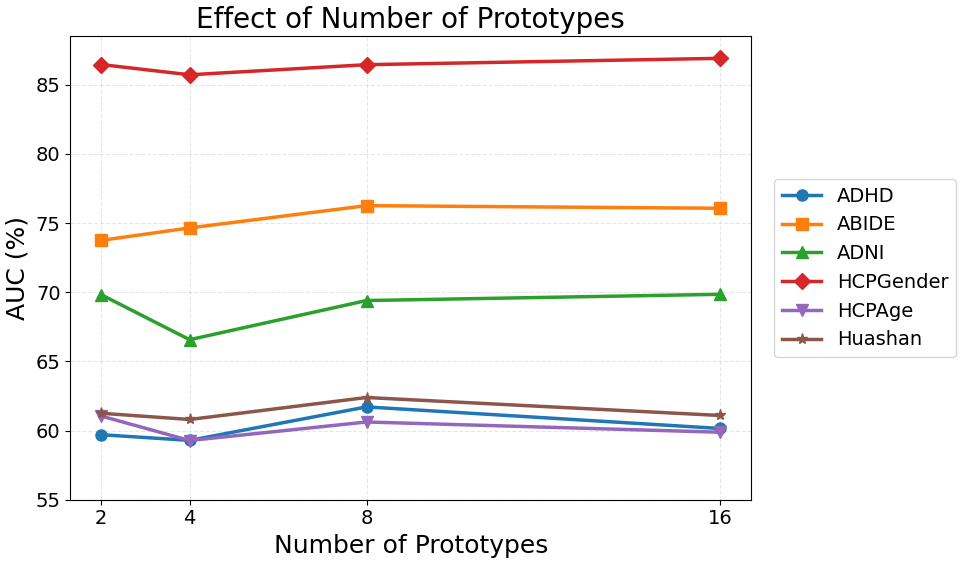}
    \label{fig:hyperparam_nproto}
}
\caption{Hyperparameter sensitivity analysis of MV-BrainFM with respect to the number of attention heads and prototypes across multiple datasets.}
\label{fig:hyperparam_analysis}
\end{figure*}

\subsection{Ablation Study}

To assess the contribution of each component in MV-BrainFM, we conduct an ablation study by progressively enabling or disabling key modules, including the Atlas Encoder, Distance Bias, and the three pretraining objectives ($\mathcal{L}_{\text{rec}}$, $\mathcal{L}_{\text{cc}}$, and $\mathcal{L}_{\text{ent}}$). All experiments are performed on networks constructed with the Schaefer100 atlas while keeping the remaining training settings unchanged.

The first row corresponds to a model that is directly fine-tuned on downstream tasks without any pretraining objectives. Although the Atlas Encoder and Distance Bias are still included in the architecture, the model is trained purely in a supervised manner. As shown in Table~\ref{tab:ablation}, this baseline achieves relatively lower performance across most datasets, indicating that the proposed pretraining strategy plays a critical role in learning transferable brain network representations.

When the three pretraining objectives ($\mathcal{L}_{\text{rec}}$, $\mathcal{L}_{\text{cc}}$, and $\mathcal{L}_{\text{ent}}$) are introduced, the performance improves consistently across datasets, even without the Atlas Encoder and Distance Bias modules. This result highlights the effectiveness of the proposed cross-view consistency learning framework in capturing shared representations across multiple atlas views.

We further observe that incorporating either the Atlas Encoder or the Distance Bias leads to additional performance improvements. The Atlas Encoder enables the model to encode spatial coordinate information and handle heterogeneous atlas configurations, while the Distance Bias introduces anatomically meaningful priors into the attention mechanism. Both modules help the model better capture the structural properties of brain networks.

Finally, when all components are jointly enabled, MV-BrainFM achieves the best performance across all datasets. This demonstrates that the architectural modules and the multi-view pretraining objectives are complementary. In particular, the combination of distance-aware modeling and cross-view consistency learning allows the model to effectively integrate anatomical priors with multi-atlas information, leading to the most robust performance.

\subsection{Hyperparameter Analysis}

We analyze the sensitivity of two key hyperparameters in MV-BrainFM: 
the number of attention heads in the DASA module and the number of prototypes $P$ used in the clustering consistency loss (Eq.~(\ref{eq:prototype})). All experiments are conducted on brain networks with the Schaefer100 atlas.

\paragraph{Number of attention heads.}
We vary the number of heads from $\{2,4,8,16\}$ and report the results in Fig.~\ref{fig:hyperparam_nhead}. 
Overall, the model performance remains relatively stable across different settings, indicating that MV-BrainFM is not overly sensitive to this hyperparameter. 
Increasing the number of heads generally improves performance from 2 to 8 heads on most datasets, suggesting that multiple heads help capture diverse connectivity patterns at different anatomical distance scales. 
However, further increasing the number of heads to 16 provides little additional benefit and sometimes slightly degrades performance, likely due to increased model complexity and over-parameterization. 
Based on these observations, we adopt 8 heads in all experiments as a good balance between performance and computational efficiency.

\paragraph{Number of prototypes.}
We further study the effect of the number of prototypes in the clustering consistency module by varying $P$ from $\{4,8,16,32\}$, as shown in Fig.~\ref{fig:hyperparam_nproto}. 
Similarly, the performance varies only moderately across different prototype numbers, demonstrating the robustness of the proposed cross-view consistency objective. 
Performance generally improves when increasing $P$ from 4 to 16, indicating that a moderate number of prototypes provides sufficient capacity to capture diverse latent patterns across views. 
When $P$ increases further to 32, the performance tends to plateau or slightly decrease on some datasets, suggesting that excessively large prototype sets may introduce unnecessary redundancy. 
Therefore, we set $P=16$ for all experiments.

Overall, these results demonstrate that MV-BrainFM maintains stable performance across a wide range of hyperparameter choices, highlighting the robustness of the proposed architecture and pretraining strategy.

\section{Conclusion}

In this work, we propose \textit{MV-BrainFM}, a multi-view brain network foundation model designed to support brain networks constructed from heterogeneous atlases. By integrating a Random Fourier Atlas Encoder, a distance-aware Transformer architecture, and a cross-view consistency pretraining strategy, the proposed framework enables unified representation learning across multiple atlas configurations while incorporating anatomical spatial priors. Extensive experiments on large-scale neuroimaging datasets demonstrate that MV-BrainFM consistently outperforms both task-specific methods and existing brain foundation models. The proposed framework also shows strong scalability with increasing pretraining data and significantly improved computational efficiency. These results highlight the potential of multi-view foundation models for scalable and interpretable brain network analysis.

Future work will focus on extending MV-BrainFM to incorporate richer sources of brain connectivity information, including additional connectivity measurements and multi-modal neuroimaging data such as structural connectivity from diffusion MRI and electrophysiological signals. Integrating multiple modalities and connectivity definitions could further improve the robustness and generalizability of learned brain representations. More broadly, scalable brain network foundation models like MV-BrainFM may facilitate large-scale neuroimaging analysis, biomarker discovery, and cross-dataset studies by enabling unified modeling across heterogeneous brain atlases and datasets. At the same time, careful attention to data privacy, dataset bias, and clinical validation will be essential to ensure responsible use of such models in neuroscience and healthcare applications.


\bibliographystyle{IEEEtrans}
\bibliography{ref}

@article{worsley2002general,
  title={A general statistical analysis for fMRI data},
  author={Worsley, Keith J and Liao, Chien Heng and Aston, John and Petre, V and Duncan, GH and Morales, F and Evans, Alan C},
  journal={Neuroimage},
  volume={15},
  number={1},
  pages={1--15},
  year={2002},
  publisher={Elsevier}
}

@inproceedings{ktena2017distance,
  title={Distance metric learning using graph convolutional networks: Application to functional brain networks},
  author={Ktena, Sofia Ira and Parisot, Sarah and Ferrante, Enzo and Rajchl, Martin and Lee, Matthew and Glocker, Ben and Rueckert, Daniel},
  booktitle={Medical Image Computing and Computer Assisted Intervention- MICCAI 2017: 20th International Conference, Quebec City, QC, Canada, September 11-13, 2017, Proceedings, Part I 20},
  pages={469--477},
  year={2017},
  organization={Springer}
}

@article{ying2018hierarchical,
  title={Hierarchical graph representation learning with differentiable pooling},
  author={Ying, Zhitao and You, Jiaxuan and Morris, Christopher and Ren, Xiang and Hamilton, Will and Leskovec, Jure},
  journal={Advances in neural information processing systems},
  volume={31},
  year={2018}
}

@inproceedings{xu2023data,
  title={Data-Driven Network Neuroscience: On Data Collection and Benchmark},
  author={Xu, Jiaxing and Yang, Yunhan and Huang, David Tse Jung and Gururajapathy, Sophi Shilpa and Ke, Yiping and Qiao, Miao and Wang, Alan and Kumar, Haribalan and McGeown, Josh and Kwon, Eryn},
  booktitle={Thirty-seventh Conference on Neural Information Processing Systems Datasets and Benchmarks Track},
  year={2023}
}

@article{kingma2014adam,
  title={Adam: A method for stochastic optimization},
  author={Kingma, Diederik P and Ba, Jimmy},
  journal={arXiv preprint arXiv:1412.6980
        
        },
  year={2014}
}

@inproceedings{kipf2017semi,
  title={Semi-Supervised Classification with Graph Convolutional Networks},
  author={Kipf, Thomas N. and Welling, Max},
  booktitle={International Conference on Learning Representations (ICLR)},
  year={2017}
}

@article{velivckovic2017graph,
  title={Graph attention networks},
  author={Veli{\v{c}}kovi{\'c}, Petar and Cucurull, Guillem and Casanova, Arantxa and Romero, Adriana and Lio, Pietro and Bengio, Yoshua},
  journal={arXiv preprint arXiv:1710.10903
        
        
        
        
        
        
        
        
        
        },
  year={2017}
}

@article{kawahara2017brainnetcnn,
  title={BrainNetCNN: Convolutional neural networks for brain networks; towards predicting neurodevelopment},
  author={Kawahara, Jeremy and Brown, Colin J and Miller, Steven P and Booth, Brian G and Chau, Vann and Grunau, Ruth E and Zwicker, Jill G and Hamarneh, Ghassan},
  journal={NeuroImage},
  volume={146},
  pages={1038--1049},
  year={2017},
  publisher={Elsevier}
}

@inproceedings{li2020pooling,
  title={Pooling regularized graph neural network for fmri biomarker analysis},
  author={Li, Xiaoxiao and Zhou, Yuan and Dvornek, Nicha C and Zhang, Muhan and Zhuang, Juntang and Ventola, Pamela and Duncan, James S},
  booktitle={Medical Image Computing and Computer Assisted Intervention--MICCAI 2020: 23rd International Conference, Lima, Peru, October 4--8, 2020, Proceedings, Part VII 23},
  pages={625--635},
  year={2020},
  organization={Springer}
}

@article{schaefer2018local,
  title={Local-global parcellation of the human cerebral cortex from intrinsic functional connectivity MRI},
  author={Schaefer, Alexander and Kong, Ru and Gordon, Evan M and Laumann, Timothy O and Zuo, Xi-Nian and Holmes, Avram J and Eickhoff, Simon B and Yeo, BT Thomas},
  journal={Cerebral cortex},
  volume={28},
  number={9},
  pages={3095--3114},
  year={2018},
  publisher={Oxford University Press}
}

@article{assaf2010abnormal,
  title={Abnormal functional connectivity of default mode sub-networks in autism spectrum disorder patients},
  author={Assaf, Michal and Jagannathan, Kanchana and Calhoun, Vince D and Miller, Laura and Stevens, Michael C and Sahl, Robert and O'Boyle, Jacqueline G and Schultz, Robert T and Pearlson, Godfrey D},
  journal={Neuroimage},
  volume={53},
  number={1},
  pages={247--256},
  year={2010},
  publisher={Elsevier}
}

@article{vaswani2017attention,
  title={Attention is all you need},
  author={Vaswani, Ashish and Shazeer, Noam and Parmar, Niki and Uszkoreit, Jakob and Jones, Llion and Gomez, Aidan N and Kaiser, {\L}ukasz and Polosukhin, Illia},
  journal={Advances in neural information processing systems},
  volume={30},
  year={2017}
}

@article{badea2017exploring,
  title={Exploring the reproducibility of functional connectivity alterations in Parkinson’s disease},
  author={Badea, Liviu and Onu, Mihaela and Wu, Tao and Roceanu, Adina and Bajenaru, Ovidiu},
  journal={PLoS One},
  volume={12},
  number={11},
  pages={e0188196},
  year={2017},
  publisher={Public Library of Science San Francisco, CA USA}
}

@article{dadi2019benchmarking,
  title={Benchmarking functional connectome-based predictive models for resting-state fMRI},
  author={Dadi, Kamalaker and Rahim, Mehdi and Abraham, Alexandre and Chyzhyk, Darya and Milham, Michael and Thirion, Bertrand and Varoquaux, Ga{\"e}l and Alzheimer's Disease Neuroimaging Initiative and others},
  journal={NeuroImage},
  volume={192},
  pages={115--134},
  year={2019},
  publisher={Elsevier}
}

@article{craddock2013neuro,
  title={The neuro bureau preprocessing initiative: open sharing of preprocessed neuroimaging data and derivatives},
  author={Craddock, Cameron and Benhajali, Yassine and Chu, Carlton and Chouinard, Francois and Evans, Alan and Jakab, Andr{\'a}s and Khundrakpam, Budhachandra Singh and Lewis, John David and Li, Qingyang and Milham, Michael and others},
  journal={Frontiers in Neuroinformatics},
  volume={7},
  pages={27},
  year={2013}
}

@article{li2021braingnn,
  title={Braingnn: Interpretable brain graph neural network for fmri analysis},
  author={Li, Xiaoxiao and Zhou, Yuan and Dvornek, Nicha and Zhang, Muhan and Gao, Siyuan and Zhuang, Juntang and Scheinost, Dustin and Staib, Lawrence H and Ventola, Pamela and Duncan, James S},
  journal={Medical Image Analysis},
  volume={74},
  pages={102233},
  year={2021},
  publisher={Elsevier}
}

@article{zhang2022classification,
  title={Classification of Brain Disorders in rs-fMRI via Local-to-Global Graph Neural Networks},
  author={Zhang, Hao and Song, Ran and Wang, Liping and Zhang, Lin and Wang, Dawei and Wang, Cong and Zhang, Wei},
  journal={IEEE Transactions on Medical Imaging},
  year={2022},
  publisher={IEEE}
}

@article{kan2022brain,
  title={Brain network transformer},
  author={Kan, Xuan and Dai, Wei and Cui, Hejie and Zhang, Zilong and Guo, Ying and Yang, Carl},
  journal={Advances in Neural Information Processing Systems},
  volume={35},
  pages={25586--25599},
  year={2022}
}

@article{xu2021graph,
  title={A graph Gaussian embedding method for predicting Alzheimer's disease progression with MEG brain networks},
  author={Xu, Mengjia and Sanz, David Lopez and Garces, Pilar and Maestu, Fernando and Li, Quanzheng and Pantazis, Dimitrios},
  journal={IEEE Transactions on Biomedical Engineering},
  volume={68},
  number={5},
  pages={1579--1588},
  year={2021},
  publisher={IEEE}
}

@article{tzourio2002automated,
  title={Automated anatomical labeling of activations in SPM using a macroscopic anatomical parcellation of the MNI MRI single-subject brain},
  author={Tzourio-Mazoyer, Nathalie and Landeau, Brigitte and Papathanassiou, Dimitri and Crivello, Fabrice and Etard, Olivier and Delcroix, Nicolas and Mazoyer, Bernard and Joliot, Marc},
  journal={Neuroimage},
  volume={15},
  number={1},
  pages={273--289},
  year={2002},
  publisher={Elsevier}
}

@article{BELLEC20101126,
title = {Multi-level bootstrap analysis of stable clusters in resting-state fMRI},
journal = {NeuroImage},
volume = {51},
number = {3},
pages = {1126-1139},
year = {2010},
issn = {1053-8119},
doi = {https://doi.org/10.1016/j.neuroimage.2010.02.082},
url = {https://www.sciencedirect.com/science/article/pii/S1053811910002697},
author = {Pierre Bellec and Pedro Rosa-Neto and Oliver C. Lyttelton and Habib Benali and Alan C. Evans}
}

@article{craddock2012whole,
  title={A whole brain fMRI atlas generated via spatially constrained spectral clustering},
  author={Craddock, R Cameron and James, G Andrew and Holtzheimer III, Paul E and Hu, Xiaoping P and Mayberg, Helen S},
  journal={Human brain mapping},
  volume={33},
  number={8},
  pages={1914--1928},
  year={2012},
  publisher={Wiley Online Library}
}

@article{xu2024contrastive,
  title={Contrastive Graph Pooling for Explainable Classification of Brain Networks},
  author={Xu, Jiaxing and Bian, Qingtian and Li, Xinhang and Zhang, Aihu and Ke, Yiping and Qiao, Miao and Zhang, Wei and Sim, Wei Khang Jeremy and Guly{\'a}s, Bal{\'a}zs},
  journal={IEEE Transactions on Medical Imaging},
  year={2024},
  publisher={IEEE}
}

@inproceedings{bannadabhavi2023community,
  title={Community-aware transformer for autism prediction in fmri connectome},
  author={Bannadabhavi, Anushree and Lee, Soojin and Deng, Wenlong and Ying, Rex and Li, Xiaoxiao},
  booktitle={International Conference on Medical Image Computing and Computer-Assisted Intervention},
  pages={287--297},
  year={2023},
  organization={Springer}
}

@article{said2023neurograph,
  title={Neurograph: Benchmarks for graph machine learning in brain connectomics},
  author={Said, Anwar and Bayrak, Roza and Derr, Tyler and Shabbir, Mudassir and Moyer, Daniel and Chang, Catie and Koutsoukos, Xenofon},
  journal={Advances in Neural Information Processing Systems},
  volume={36},
  pages={6509--6531},
  year={2023}
}

@inproceedings{xu2024contrasformer,
  title={Contrasformer: A Brain Network Contrastive Transformer for Neurodegenerative Condition Identification},
  author={Xu, Jiaxing and He, Kai and Lan, Mengcheng and Bian, Qingtian and Li, Wei and Li, Tieying and Ke, Yiping and Qiao, Miao},
  booktitle={Proceedings of the 33rd ACM International Conference on Information and Knowledge Management},
  pages={2671--2681},
  year={2024}
}

@article{marek2011parkinson,
  title={The Parkinson progression marker initiative (PPMI)},
  author={Marek, Kenneth and Jennings, Danna and Lasch, Shirley and Siderowf, Andrew and Tanner, Caroline and Simuni, Tanya and Coffey, Chris and Kieburtz, Karl and Flagg, Emily and Chowdhury, Sohini and others},
  journal={Progress in neurobiology},
  volume={95},
  number={4},
  pages={629--635},
  year={2011},
  publisher={Elsevier}
}

@inproceedings{
peng2025biologically,
title={Biologically Plausible Brain Graph Transformer},
author={Ciyuan Peng and Yuelong Huang and Qichao Dong and Shuo Yu and Feng Xia and Chengqi Zhang and Yaochu Jin},
booktitle={The Thirteenth International Conference on Learning Representations},
year={2025},
url={https://openreview.net/forum?id=rQyg6MnsDb}
}

@article{shehzad2025multiscale,
  title={Multiscale graph transformer for brain disorder diagnosis},
  author={Shehzad, Ahsan and Zhang, Dongyu and Yu, Shuo and Abid, Shagufta and Kumar, Dinesh Kant and Xia, Feng},
  journal={IEEE Transactions on Consumer Electronics},
  year={2025},
  publisher={IEEE}
}

@inproceedings{yang2023ptgb,
  title={PTGB: Pre-Train Graph Neural Networks for Brain Network Analysis},
  author={Yang, Yi and Cui, Hejie and Yang, Carl},
  booktitle={Conference on Health, Inference, and Learning},
  pages={526--544},
  year={2023},
  organization={PMLR}
}

@article{yang2024brainmass,
  title={Brainmass: Advancing brain network analysis for diagnosis with large-scale self-supervised learning},
  author={Yang, Yanwu and Ye, Chenfei and Su, Guinan and Zhang, Ziyao and Chang, Zhikai and Chen, Hairui and Chan, Piu and Yu, Yue and Ma, Ting},
  journal={IEEE transactions on medical imaging},
  volume={43},
  number={11},
  pages={4004--4016},
  year={2024},
  publisher={IEEE}
}

@article{dong2024brain,
  title={Brain-jepa: Brain dynamics foundation model with gradient positioning and spatiotemporal masking},
  author={Dong, Zijian and Li, Ruilin and Wu, Yilei and Nguyen, Thuan Tinh and Chong, Joanna and Ji, Fang and Tong, Nathanael and Chen, Christopher and Zhou, Juan Helen},
  journal={Advances in Neural Information Processing Systems},
  volume={37},
  pages={86048--86073},
  year={2024}
}

@inproceedings{yangwe,
  title={Do We Really Need Message Passing in Brain Network Modeling?},
  author={Yang, Liang and Liu, Yuwei and Zhuo, Jiaming and Jin, Di and Wang, Chuan and Wang, Zhen and Cao, Xiaochun},
  year={2025},
  booktitle={Forty-second International Conference on Machine Learning}
}

@article{xu2025multi,
  title={Multi-atlas brain network classification through consistency distillation and complementary information fusion},
  author={Xu, Jiaxing and Lan, Mengcheng and Dong, Xia and He, Kai and Zhang, Wei and Bian, Qingtian and Ke, Yiping},
  journal={IEEE Journal of Biomedical and Health Informatics},
  year={2025},
  publisher={IEEE}
}

@article{radford2018improving,
  title={Improving language understanding by generative pre-training},
  author={Radford, Alec and Narasimhan, Karthik and Salimans, Tim and Sutskever, Ilya and others},
  year={2018},
  publisher={San Francisco, CA, USA}
}

@article{dosovitskiy2020image,
  title={An image is worth 16x16 words: Transformers for image recognition at scale},
  author={Dosovitskiy, Alexey},
  journal={arXiv preprint arXiv:2010.11929},
  year={2020}
}

@article{van2013wu,
  title={The WU-Minn human connectome project: an overview},
  author={Van Essen, David C and Smith, Stephen M and Barch, Deanna M and Behrens, Timothy EJ and Yacoub, Essa and Ugurbil, Kamil and Wu-Minn HCP Consortium and others},
  journal={Neuroimage},
  volume={80},
  pages={62--79},
  year={2013},
  publisher={Elsevier}
}

@article{wang2026graph,
  title={A graph transformer-based foundation model for brain functional connectivity network},
  author={Wang, Yulong and Calhoun, Vince D and Pearlson, Godfrey D and Kochunov, Peter and van Erp, Theo GM and Du, Yuhui},
  journal={Pattern Recognition},
  volume={169},
  pages={111988},
  year={2026},
  publisher={Elsevier}
}

@article{ding2016progression,
  title={Progression and predictors of mild cognitive impairment in Chinese elderly: a prospective follow-up in the Shanghai Aging Study},
  author={Ding, Ding and Zhao, Qianhua and Guo, Qihao and Liang, Xiaoniu and Luo, Jianfeng and Yu, Lirong and Zheng, Li and Hong, Zhen and SAS, Shanghai Aging Study},
  journal={Alzheimer's \& Dementia: Diagnosis, Assessment \& Disease Monitoring},
  volume={4},
  pages={28--36},
  year={2016},
  publisher={Elsevier}
}

@article{liu2022multiscale,
  title={Multiscale functional connectome abnormality predicts cognitive outcomes in subcortical ischemic vascular disease},
  author={Liu, Mianxin and Wang, Yao and Zhang, Han and Yang, Qing and Shi, Feng and Zhou, Yan and Shen, Dinggang},
  journal={Cerebral Cortex},
  volume={32},
  number={21},
  pages={4641--4656},
  year={2022},
  publisher={Oxford Academic}
}

@article{marcus2007open,
  title={Open Access Series of Imaging Studies (OASIS): cross-sectional MRI data in young, middle aged, nondemented, and demented older adults},
  author={Marcus, Daniel S and Wang, Tracy H and Parker, Jamie and Csernansky, John G and Morris, John C and Buckner, Randy L},
  journal={Journal of cognitive neuroscience},
  volume={19},
  number={9},
  pages={1498--1507},
  year={2007},
  publisher={MIT Press One Rogers Street, Cambridge, MA 02142-1209, USA journals-info~…}
}

@article{adhd2012adhd,
  title={The ADHD-200 consortium: a model to advance the translational potential of neuroimaging in clinical neuroscience},
  author={ADHD-200 consortium},
  journal={Frontiers in systems neuroscience},
  volume={6},
  pages={62},
  year={2012},
  publisher={Frontiers Research Foundation}
}

@article{yan2019reduced,
  title={Reduced default mode network functional connectivity in patients with recurrent major depressive disorder},
  author={Yan, Chao-Gan and Chen, Xiao and Li, Le and Castellanos, Francisco Xavier and Bai, Tong-Jian and Bo, Qi-Jing and Cao, Jun and Chen, Guan-Mao and Chen, Ning-Xuan and Chen, Wei and others},
  journal={Proceedings of the National Academy of Sciences},
  volume={116},
  number={18},
  pages={9078--9083},
  year={2019},
  publisher={National Academy of Sciences}
}

@article{alexander2017open,
  title={An open resource for transdiagnostic research in pediatric mental health and learning disorders},
  author={Alexander, Lindsay M and Escalera, Jasmine and Ai, Lei and Andreotti, Charissa and Febre, Karina and Mangone, Alexander and Vega-Potler, Natan and Langer, Nicolas and Alexander, Alexis and Kovacs, Meagan and others},
  journal={Scientific data},
  volume={4},
  number={1},
  pages={1--26},
  year={2017},
  publisher={Nature Publishing Group}
}

@article{liu2023common,
  title={A common spectrum underlying brain disorders across lifespan revealed by deep learning on brain networks},
  author={Liu, Mianxin and Zhang, Jingyang and Wang, Yao and Zhou, Yan and Xie, Fang and Guo, Qihao and Shi, Feng and Zhang, Han and Wang, Qian and Shen, Dinggang},
  journal={Iscience},
  volume={26},
  number={11},
  year={2023},
  publisher={Elsevier}
}

@article{wang2015gretna,
title={GRETNA: a graph theoretical network analysis toolbox for imaging connectomics},
author={Wang, Jinhui and Wang, Xindi and Xia, Mingrui and Liao, Xuhong and Evans, Alan and He, Yong},
journal={Frontiers in human neuroscience},
volume={9},
pages={386},
year={2015},
publisher={Frontiers Media SA}
}

@article{ercsey2013predictive,
  title={A predictive network model of cerebral cortical connectivity based on a distance rule},
  author={Ercsey-Ravasz, M{\'a}ria and Markov, Nikola T and Lamy, Camille and Van Essen, David C and Knoblauch, Kenneth and Toroczkai, Zolt{\'a}n and Kennedy, Henry},
  journal={Neuron},
  volume={80},
  number={1},
  pages={184--197},
  year={2013},
  publisher={Elsevier}
}

@article{goni2014resting,
  title={Resting-brain functional connectivity predicted by analytic measures of network communication},
  author={Go{\~n}i, Joaqu{\'\i}n and Van Den Heuvel, Martijn P and Avena-Koenigsberger, Andrea and Velez de Mendizabal, Nieves and Betzel, Richard F and Griffa, Alessandra and Hagmann, Patric and Corominas-Murtra, Bernat and Thiran, Jean-Philippe and Sporns, Olaf},
  journal={Proceedings of the National Academy of Sciences},
  volume={111},
  number={2},
  pages={833--838},
  year={2014},
  publisher={National Academy of Sciences}
}

@article{vertes2012simple,
  title={Simple models of human brain functional networks},
  author={V{\'e}rtes, Petra E and Alexander-Bloch, Aaron F and Gogtay, Nitin and Giedd, Jay N and Rapoport, Judith L and Bullmore, Edward T},
  journal={Proceedings of the National Academy of Sciences},
  volume={109},
  number={15},
  pages={5868--5873},
  year={2012},
  publisher={National Academy of Sciences}
}

@inproceedings{
wei2026a,
title={A Brain Graph Foundation Model: Pre-Training and Prompt-Tuning across Broad Atlases and Disorders},
author={Xinxu Wei and kanhao zhao and Yong Jiao and Lifang He and Yu Zhang},
booktitle={The Fourteenth International Conference on Learning Representations},
year={2026},
url={https://openreview.net/forum?id=PeGHkAaRxs}
}

@article{bullmore2012economy,
  title={The economy of brain network organization},
  author={Bullmore, Ed and Sporns, Olaf},
  journal={Nature reviews neuroscience},
  volume={13},
  number={5},
  pages={336--349},
  year={2012},
  publisher={Nature Publishing Group UK London}
}

@article{abraham2014machine,
  title={Machine learning for neuroimaging with scikit-learn},
  author={Abraham, Alexandre and Pedregosa, Fabian and Eickenberg, Michael and Gervais, Philippe and Mueller, Andreas and Kossaifi, Jean and Gramfort, Alexandre and Thirion, Bertrand and Varoquaux, Ga{\"e}l},
  journal={Frontiers in neuroinformatics},
  volume={8},
  pages={14},
  year={2014},
  publisher={Frontiers Media SA}
}

@article{wang2006neural,
  title={Neural basis of irony comprehension in children with autism: the role of prosody and context},
  author={Wang, A Ting and Lee, Susan S and Sigman, Marian and Dapretto, Mirella},
  journal={Brain},
  volume={129},
  number={4},
  pages={932--943},
  year={2006},
  publisher={Oxford University Press}
}

@article{soulieres2009enhanced,
  title={Enhanced visual processing contributes to matrix reasoning in autism},
  author={Souli{\`e}res, Isabelle and Dawson, Michelle and Samson, Fabienne and Barbeau, Elise B and Sahyoun, Cherif P and Strangman, Gary E and Zeffiro, Thomas A and Mottron, Laurent},
  journal={Human brain mapping},
  volume={30},
  number={12},
  pages={4082--4107},
  year={2009},
  publisher={Wiley Online Library}
}

@article{damoiseaux2012functional,
  title={Functional connectivity tracks clinical deterioration in Alzheimer's disease},
  author={Damoiseaux, Jessica S and Prater, Katherine E and Miller, Bruce L and Greicius, Michael D},
  journal={Neurobiology of aging},
  volume={33},
  number={4},
  pages={828--e19},
  year={2012},
  publisher={Elsevier}
}

@article{bosch2010cognitive,
  title={Cognitive reserve modulates task-induced activations and deactivations in healthy elders, amnestic mild cognitive impairment and mild Alzheimer's disease},
  author={Bosch, Beatriz and Bartr{\'e}s-Faz, David and Rami, Lorena and Arenaza-Urquijo, Eider M and Fern{\'a}ndez-Espejo, Davinia and Junqu{\'e}, Carme and Sol{\'e}-Padull{\'e}s, Cristina and Sanchez-Valle, Raquel and Bargallo, Nuria and Falcon, Carles and others},
  journal={Cortex},
  volume={46},
  number={4},
  pages={451--461},
  year={2010},
  publisher={Elsevier}
}

@article{ma2016ventral,
  title={Ventral striatal hyperconnectivity during rewarded interference control in adolescents with ADHD},
  author={Ma, Ili and van Holstein, Mieke and Mies, Gabry W and Mennes, Maarten and Buitelaar, Jan and Cools, Roshan and Cillessen, Antonius HN and Krebs, Ruth M and Scheres, Anouk},
  journal={Cortex},
  volume={82},
  pages={225--236},
  year={2016},
  publisher={Elsevier}
}

@article{o2018neural,
  title={Neural circuitry underlying sustained attention in healthy adolescents and in ADHD symptomatology},
  author={O'Halloran, Laura and Cao, Zhipeng and Ruddy, Kathy and Jollans, Lee and Albaugh, Matthew D and Aleni, Andrea and Potter, Alexandra S and Vahey, Nigel and Banaschewski, Tobias and Hohmann, Sarah and others},
  journal={Neuroimage},
  volume={169},
  pages={395--406},
  year={2018},
  publisher={Elsevier}
}

@article{deng2016mapping,
  title={Mapping the “what” and “where” visual cortices and their atrophy in Alzheimer's disease: Combined activation likelihood estimation with voxel-based morphometry},
  author={Deng, Yanjia and Shi, Lin and Lei, Yi and Liang, Peipeng and Li, Kuncheng and Chu, Winnie CW and Wang, Defeng and Alzheimer's Disease Neuroimaging Initiative},
  journal={Frontiers in Human Neuroscience},
  volume={10},
  pages={333},
  year={2016},
  publisher={Frontiers Media SA}
}

@inproceedings{
caro2024brainlm,
title={Brain{LM}: A foundation model for brain activity recordings},
author={Josue Ortega Caro and Antonio Henrique de Oliveira Fonseca and Syed A Rizvi and Matteo Rosati and Christopher Averill and James L Cross and Prateek Mittal and Emanuele Zappala and Rahul Madhav Dhodapkar and Chadi Abdallah and David van Dijk},
booktitle={The Twelfth International Conference on Learning Representations},
year={2024},
url={https://openreview.net/forum?id=RwI7ZEfR27}
}

@article{xu2025multiview,
  title={Multi-View Graph Learning in Brain Network Analysis: A Survey},
  author={Xu, Jiaxing and Dong, Xia and Huang, Tiancheng and Wu, Danyang and Zhang, Wei and Ke, Yiping and Yu, Philip S},
  journal={Authorea Preprints},
  year={2025},
  publisher={Authorea}
}

@article{arslan2018human,
  title={Human brain mapping: A systematic comparison of parcellation methods for the human cerebral cortex},
  author={Arslan, Salim and Ktena, Sofia Ira and Makropoulos, Antonios and Robinson, Emma C and Rueckert, Daniel and Parisot, Sarah},
  journal={NeuroImage},
  volume={170},
  pages={5--30},
  year={2018},
  publisher={Elsevier}
}

@article{parisot2018disease,
  title={Disease prediction using graph convolutional networks: application to autism spectrum disorder and Alzheimer’s disease},
  author={Parisot, Sarah and Ktena, Sofia Ira and Ferrante, Enzo and Lee, Matthew and Guerrero, Ricardo and Glocker, Ben and Rueckert, Daniel},
  journal={Medical image analysis},
  volume={48},
  pages={117--130},
  year={2018},
  publisher={Elsevier}
}

@article{hu2021gat,
  title={GAT-LI: a graph attention network based learning and interpreting method for functional brain network classification},
  author={Hu, Jinlong and Cao, Lijie and Li, Tenghui and Dong, Shoubin and Li, Ping},
  journal={BMC bioinformatics},
  volume={22},
  number={1},
  pages={379},
  year={2021},
  publisher={Springer}
}

@article{ying2019gnnexplainer,
  title={Gnnexplainer: Generating explanations for graph neural networks},
  author={Ying, Zhitao and Bourgeois, Dylan and You, Jiaxuan and Zitnik, Marinka and Leskovec, Jure},
  journal={Advances in neural information processing systems},
  volume={32},
  year={2019}
}

@inproceedings{
hu2020strategies,
title={Strategies for Pre-training Graph Neural Networks},
author={Weihua Hu* and Bowen Liu* and Joseph Gomes and Marinka Zitnik and Percy Liang and Vijay Pande and Jure Leskovec},
booktitle={International Conference on Learning Representations},
year={2020},
url={https://openreview.net/forum?id=HJlWWJSFDH}
}

@inproceedings{hou2022graphmae,
  title={Graphmae: Self-supervised masked graph autoencoders},
  author={Hou, Zhenyu and Liu, Xiao and Cen, Yukuo and Dong, Yuxiao and Yang, Hongxia and Wang, Chunjie and Tang, Jie},
  booktitle={Proceedings of the 28th ACM SIGKDD conference on knowledge discovery and data mining},
  pages={594--604},
  year={2022}
}

@article{liu2025graph,
  title={Graph foundation models: Concepts, opportunities and challenges},
  author={Liu, Jiawei and Yang, Cheng and Lu, Zhiyuan and Chen, Junze and Li, Yibo and Zhang, Mengmei and Bai, Ting and Fang, Yuan and Sun, Lichao and Yu, Philip S and others},
  journal={IEEE Transactions on Pattern Analysis and Machine Intelligence},
  year={2025},
  publisher={IEEE}
}

@article{liu2023deep,
  title={Deep fusion of multi-template using spatio-temporal weighted multi-hypergraph convolutional networks for brain disease analysis},
  author={Liu, Jingyu and Cui, Weigang and Chen, Yipeng and Ma, Yulan and Dong, Qunxi and Cai, Ran and Li, Yang and Hu, Bin},
  journal={IEEE Transactions on Medical Imaging},
  volume={43},
  number={2},
  pages={860--873},
  year={2023},
  publisher={IEEE}
}

@article{wang2024multiview,
  title={Multiview hyperedge-aware hypergraph embedding learning for multisite, multiatlas fMRI based functional connectivity network analysis},
  author={Wang, Wei and Xiao, Li and Qu, Gang and Calhoun, Vince D and Wang, Yu-Ping and Sun, Xiaoyan},
  journal={Medical Image Analysis},
  volume={94},
  pages={103144},
  year={2024},
  publisher={Elsevier}
}

@article{yun2019graph,
  title={Graph transformer networks},
  author={Yun, Seongjun and Jeong, Minbyul and Kim, Raehyun and Kang, Jaewoo and Kim, Hyunwoo J},
  journal={Advances in neural information processing systems},
  volume={32},
  year={2019}
}

@inproceedings{
jiang2024large,
title={Large Brain Model for Learning Generic Representations with Tremendous {EEG} Data in {BCI}},
author={Wei-Bang Jiang and Li-Ming Zhao and Bao-Liang Lu},
booktitle={The Twelfth International Conference on Learning Representations},
year={2024},
url={https://openreview.net/forum?id=QzTpTRVtrP}
}

@article{banville2025scaling,
  title={Scaling laws for decoding images from brain activity},
  author={Banville, Hubert and Benchetrit, Yohann and d'Ascoli, St{\'e}phane and Rapin, J{\'e}r{\'e}my and King, Jean-R{\'e}mi},
  journal={arXiv preprint arXiv:2501.15322},
  year={2025}
}

@article{cooper2017reduced,
  title={Reduced hippocampal functional connectivity during episodic memory retrieval in autism},
  author={Cooper, Rose A and Richter, Franziska R and Bays, Paul M and Plaisted-Grant, Kate C and Baron-Cohen, Simon and Simons, Jon S},
  journal={Cerebral Cortex},
  volume={27},
  number={2},
  pages={888--902},
  year={2017},
  publisher={Oxford University Press}
}

@inproceedings{mahler2023pretraining,
  title={Pretraining is All You Need: A Multi-Atlas Enhanced Transformer Framework for Autism Spectrum Disorder Classification},
  author={Mahler, Lucas and Wang, Qi and Steiglechner, Julius and Birk, Florian and Heczko, Samuel and Scheffler, Klaus and Lohmann, Gabriele},
  booktitle={International Workshop on Machine Learning in Clinical Neuroimaging},
  pages={123--132},
  year={2023},
  organization={Springer}
}

@article{fonov2009unbiased,
  title={Unbiased nonlinear average age-appropriate brain templates from birth to adulthood},
  author={Fonov, Vladimir S and Evans, Alan C and McKinstry, Robert C and Almli, C Robert and Collins, DL},
  journal={NeuroImage},
  volume={47},
  pages={S102},
  year={2009},
  publisher={Elsevier}
}

\vfill

\end{document}